\documentclass[10pt,twocolumn,letterpaper]{article}

\usepackage{cvpr}              
\usepackage{multirow}
\usepackage{mathtools}

\usepackage[table]{xcolor}
\newcommand{\oursbg}{\cellcolor{gray!10}}

\newcommand{\tablestyle}[2]{\setlength{\tabcolsep}{#1}\renewcommand{\arraystretch}{#2}\centering\footnotesize}
\usepackage{makecell}
\usepackage{multicol}
\usepackage{multirow}
\usepackage{xcolor}
\usepackage{colortbl}
\definecolor{gr}{gray}{0.95}

\definecolor{convcolor}{HTML}{412F8A}
\definecolor{vitcolor}{HTML}{fc8e62}
\newcommand{\convcolor}[1]{\textcolor{convcolor}{#1}}
\newcommand{\vitcolor}[1]{\textcolor{vitcolor}{#1}}
\newcommand{\vb}{\vitcolor{$\mathbf{\circ}$\,}}
\newcommand{\cb}{\convcolor{$\bullet$\,}}
\definecolor{cvprblue}{rgb}{0.21,0.49,0.74}
\usepackage[pagebackref,breaklinks,colorlinks,allcolors=cvprblue]{hyperref}

\def\paperID{41268} 
\def\confName{CVPR}
\def\confYear{2026}

\title{Cross-modal Identity Mapping: Minimizing Information Loss in Modality Conversion via Reinforcement Learning}

\author{
Haonan Jia\textsuperscript{1}\thanks{Equal contribution.}, 
Shichao Dong\textsuperscript{1}\footnotemark[1], 
Xin Dong\textsuperscript{1}, 
Zenghui Sun\textsuperscript{1}, 
Jin Wang\textsuperscript{2}, \\
Jinsong Lan\textsuperscript{1}, 
Xiaoyong Zhu\textsuperscript{1}, 
Bo Zheng\textsuperscript{1}\thanks{Corresponding author.}, 
Kaifu Zhang\textsuperscript{1}\\
\textsuperscript{1}Taobao \& Tmall Group of Alibaba \quad 
\textsuperscript{2}The University of Hong Kong\\
{\tt\small 
\{haonanjia11,dongshichao1996,ddjyhhd\}@gmail.com,
\{wj0529\}@connect.hku.hk,
}\\
{\tt\small 
\{zenghui.szh,jinsonglan.ljs\}@taobao.com,
\{xiaoyong.z,bozheng\}@alibaba-inc.com
}
}

\begin{document}
\maketitle
\begin{abstract}
Large Vision-Language Models (LVLMs) often omit or misrepresent critical visual content in generated image captions. 
Minimizing such information loss will force LVLMs to focus on image details to generate precise descriptions. 
However, measuring information loss during modality conversion is inherently challenging due to the modal gap between visual content and text output. 
In this paper, we argue that the quality of an image caption is positively correlated with the similarity between images retrieved via text search using that caption. 
Based on this insight, we further propose Cross-modal Identity Mapping (CIM), a reinforcement learning framework that enhances image captioning without requiring additional annotations. 
Specifically, the method quantitatively evaluates the information loss from two perspectives: Gallery Representation Consistency and Query-gallery Image Relevance. 
Supervised under these metrics, LVLM minimizes information loss and aims to achieve identity mapping from images to captions. 
The experimental results demonstrate the superior performance of our method in image captioning, even when compared with Supervised Fine-Tuning. 
Particularly, on the COCO-LN500 benchmark, CIM achieves a 20\% improvement in relation reasoning on Qwen2.5-VL-7B. 
The code is released on \href{https://github.com/Jia-hn/Cross-modal-Identity-Mapping}{GitHub}. 
\end{abstract}
\section{Introduction}
\label{sec:intro}

Image captioning~\cite{rotstein2024fusecap,huang2019attention,liu2017improved} plays an essential role in a wide range of multimodal understanding tasks, e.g., visual question answering~\cite{cheng2025simplevqa,hu2024omnimedvqa}, image-text retrieval~\cite{duan2025fuzzy,chen2024make}, and compositional reasoning~\cite{saravanan2025velociti,qiu2025step}. 
Recently, Large Vision-Language Models (LVLMs)~\cite{instructblip,liu2023improvedllava,dong2024internlm,bai2023qwen,ye2023mplugowl} have achieved significant strides and attracted widespread attention in this task.
However, such LVLM-based methods~\cite{cornia2020meshed, huang2019attention, liu2017attention, liu2017improved,feng2019unsupervised, bahng2025cycle,tewel2022zerocap, xu2023zero} struggle to generate fine-grained captions and often misrepresent critical visual details in their textual outputs. 

\begin{figure}[t!]
    \centering
    \includegraphics[width=\linewidth]{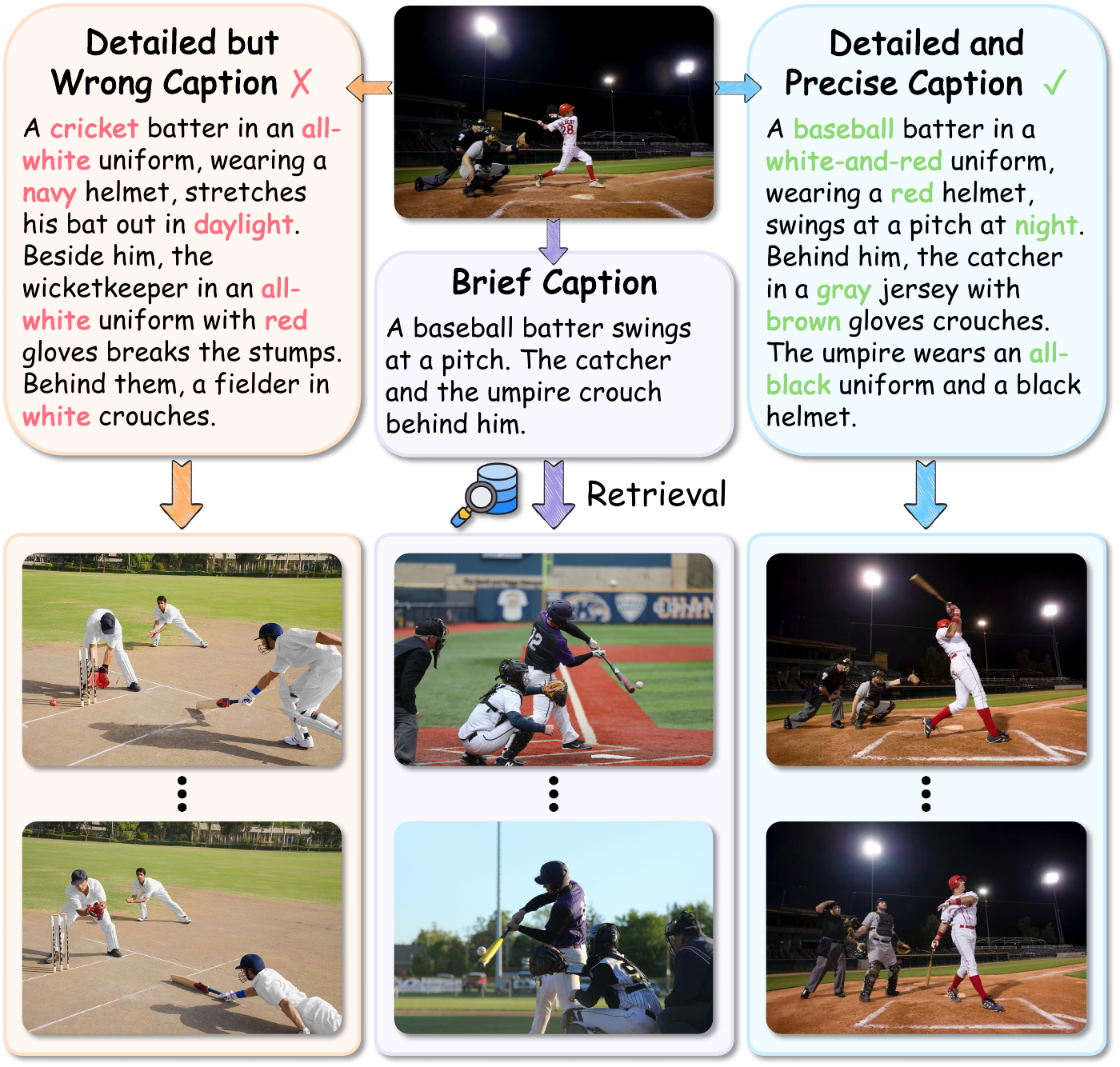}
    \caption{ 
    \textbf{Qualitative comparison of retrieved images using different captions. }
    The more detailed the caption, the higher the semantic consistency among the retrieved images. 
The more accurate the caption, the greater the visual similarity between the retrieved images and the source image. 
    }
    \label{fig:problem}
\end{figure}

Previous studies on improving LVLMs for image captioning have primarily focused on two perspectives. 
Several studies~\cite{luo2024unleashing, yang2025bacon, li2024densefusion, chen2024sharegpt4v} construct datasets with detailed image descriptions to fine-tune LVLMs to generate fine-grained captions. 
However, creating such datasets requires costly and time-consuming human annotation. 
Especially, the data construction burden will be pronounced when fine-grained visual details must be meticulously labeled at scale.
Other methods present metrics to quantify the information loss between the image and its corresponding captions. 
By using metrics as reward functions in reinforcement learning, these methods force LVLMs to generate fine-grained captions.
For instance, SC-Captioner \cite{zhang2025sc} evaluates the completeness of the caption by checking whether it contains predefined words.
However, this approach lacks explicit supervision for non-keyword elements.
Some metrics~\cite{cho2022fine, yu2023cgt, dzabraev2024vlrm,liu2018show, gaur2024no, dessi2023cross} analyze information loss in image captions based on Vision Language Models (VLMs).
Unfortunately, these methods may exhibit reward hacking in reinforcement learning, due to the limited compositional reasoning ability of VLMs~\cite{yuksekgonul2022and,wang2024diagnosing}. 
Consequently, accurately quantifying the information loss in image captioning remains a challenging and underexplored problem.

We argue that the more fine-grained the caption for the source image, the higher the representation similarity among images retrieved using that caption.
In parallel, more accurate captions will lead to higher relevance between retrieved images and the source image. 
Specifically, we enumerate three possible scenarios for image captions and perform image retrieval.
As shown in \cref{fig:problem},
when captions remain inconsistent with visual contents,  the retrieved results demonstrate significantly reduced relevance to the source image.
In contrast, images retrieved based on correct captions are similar to the source image.
Notably, retrieved images exhibit higher consistency in representation when the query text captures fine-grained details from the source image.
In this way, we can infer the information loss between the image and caption by analyzing the distribution of images retrieved by this caption.

Based on this insight, we propose Cross-modal Identity Mapping (CIM), an annotation-free reinforcement learning framework.
This method quantifies information loss in modality conversion from two perspectives: Gallery Representation Consistency (GRC) and Query-gallery Image Relevance (QIR). 
Specifically, the GRC metric evaluates the consistency among retrieved images as an indicator of the detail richness in image captions.
Additionally, Query-gallery Image Relevance measures caption precision by analyzing the similarity between the source image and retrieved images.
Through these two metrics, we quantitatively assess the precision and detail richness of the caption, thereby evaluating the information loss in image-to-text generation.
Furthermore, we formulate these two metrics as a reward signal and optimize LVLMs via reinforcement learning.
In this way, the model reduces information loss during image captioning and aims to achieve identity mapping from images to captions. 
In experiments, we validate the effectiveness of our approach across various models. 
Our contributions can be summarized as follows:
\begin{itemize}
    \item
    We convert the quantification of image caption quality into a measure of image-to-image similarity.
    In this way, we introduce metrics to quantitatively evaluate the information loss in modality conversion from two perspectives, without requiring any human annotations. 
    \item We propose Cross-modal Identity Mapping (CIM) that uses proposed metrics as rewards to encourage LVLMs to generate fine-grained and precise captions. 
    \item Extensive experiments demonstrate that our method improved the performance on various LVLMs, outperforming state-of-the-art methods by a large margin. 
\end{itemize}

\section{Related Work}

\begin{figure*}[t]
    \centering
    \includegraphics[width=\textwidth]{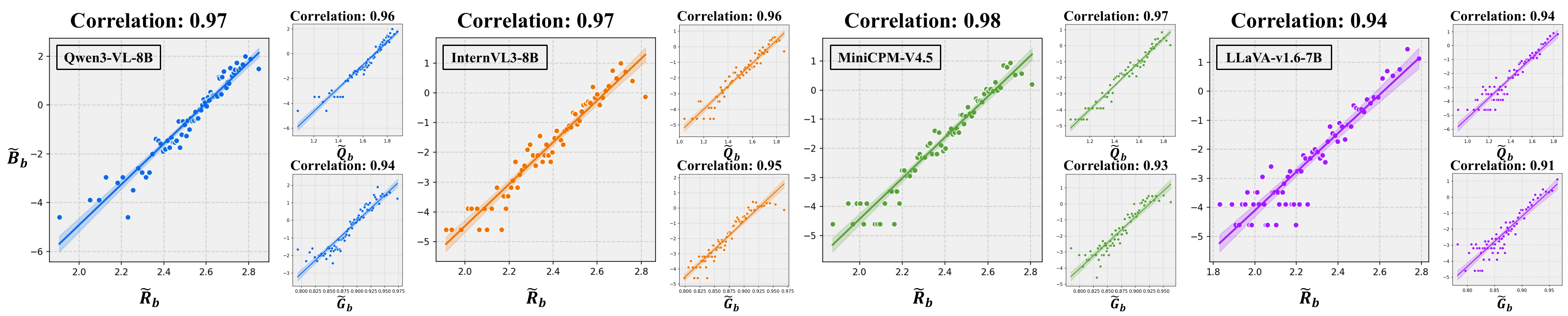}
    \caption{\textbf{Verifying the effectiveness of our proposed metrics for quantifying information loss via Pearson correlation.} For each LVLM, we plot the correlation between average metrics and the logit of breed classification accuracy. Each point corresponds to one bin, with each bin containing 100 samples. The Pearson correlation coefficient is reported for each model. This correlation verifies the effectiveness of our proposed metrics for quantifying information loss during modality conversion.}
    \label{fig:correlation}
\end{figure*}

Image captioning is a fundamental task that bridges the modality gap between vision and language, translating visual inputs into semantically grounded natural-language descriptions.
Recent work primarily focuses on generating fine-grained captions while mitigating hallucinations, broadly along three lines: supervised fine-tuning (SFT), reinforcement learning (RL), and training-free.
\subsection{Supervised Fine-Tuning}
Supervised fine-tuning leverages detailed annotated image–caption pairs to provide explicit supervision, enabling models to generate accurate and comprehensive image captions.
Previous methods solve this task primarily by using an encoder to encode the image and a decoder to generate the caption~\cite{cornia2020meshed, huang2019attention, liu2017attention, liu2017improved, vinyals2015show, wang2022git, mokady2021clipcap, luo2023tuning}.
Building on this, some methods employ retrieval-augmented generation (RAG) to condition caption generation on the image and text retrieved from external corpora~\cite{ramos2023smallcap, li2024evcap, kim2025vipcap}.
Complementary approaches substitute human annotations with synthetic data for supervision~\cite{luo2024unleashing, yang2025bacon, li2024densefusion, chen2024sharegpt4v}.
In addition, some approaches leverage the shared embedding space learned by contrastively trained vision–language encoders for self-supervised training~\cite{fei2023transferable, tam2023simple, lee2025diffusion}.
Beyond monolithic decoders, a panoptic pipeline decomposes the task into multiple simpler subtasks~\cite{lin2025panoptic}.
Controllability is introduced by fine-tuning captioners to follow user-specified controls~\cite{kornblith2023guiding, saito2025captionsmiths}.
Although these approaches substantially enhance the accuracy and comprehensiveness of generated captions, they generally rely on large-scale, detailed image–caption datasets, whose collection and annotation are costly and time-consuming.

\subsection{Reinforcement Learning}
In addition, some methods adopt reinforcement learning to optimize image captioning performance in LVLMs through task-specific reward signals. 
One representative approach leverages CLIP-based image–text similarity~\cite{radford2021learning} to score the alignment between an image and generated caption~\cite{cho2022fine, yu2023cgt, dzabraev2024vlrm}.
Complementary approaches enforce cycle consistency by reconstructing the image from the caption and rewarding captions that enable faithful regeneration~\cite{feng2019unsupervised, bahng2025cycle}. 
In addition, some approaches adopt a self-retrieval reward signal, rewarding captions that can retrieve the source image~\cite{liu2018show, gaur2024no, dessi2023cross}.
Furthermore, reinforcement learning is also employed to encourage models to self-correct~\cite{zhang2025sc}.
More recent work uses downstream task performance, conditioned on the generated caption, as the reward signal~\cite{xing2025caprl, miao2026seeing}.

\subsection{Training-Free}
There are also many training-free approaches for image captioning tasks. 
One line of work uses CLIP at inference time to guide sampling, steering the model toward captions that are better aligned with the image~\cite{tewel2022zerocap, xu2023zero}.
Another line scales test-time compute to further strengthen caption quality without any additional training~\cite{zeng2024meacap, pi2024image, peng2025patch, jung2025visual, lee2024toward,  deria2025dual, zhao2025mitigating}. 
Beyond these, a third class of methods also requires no training~\cite{leng2023mitigating, huang2024opera, favero2024multi, chen2024halc,park2024convis,dong2025inter,zou2024look,an2024agla,qu2024look}. 
Instead, they employ contrastive decoding during inference to progressively enhance the model’s focus on salient image content. 
However, these approaches all introduce additional computational overhead during inference, which hinders their practical deployment. 
This motivates the need for methods that encourage the model during training to attend to richer visual details and generate precise captions. 

\section{Method}
In this section, we expect to quantitatively evaluate the information loss in image-to-text conversion during image captioning.
Specifically, we first verify the existence of information loss across various Large Vision Language Models (LVLMs).
Second, we present a quantitative evaluation metric to assess the quality of LVLM-generated image captions.
In the verification experiment, we demonstrate the positive correlation between the metric and the quality of an image caption.
Finally, we introduce Cross-modal Identity Mapping, a novel semi-supervised method to improve model performance in image captioning.

\subsection{Verifying the Existence of Information Loss}
\label{sec:3.1}
LVLMs struggle to generate comprehensive and accurate descriptions of images, leading to critical visual details being omitted or misrepresented in the textual output. 
We conduct a fine-grained image classification experiment across various LVLMs to verify this phenomenon.

Fine-grained categories can represent more nuanced differences in visual content.
In other words, when LVLM-generated captions incorporate fine-grained category descriptions rather than broad pronouns, it demonstrates that the model captures more visual details during image captioning.
To this end, we evaluate the information loss between images and LVLM-generated captions on the Oxford-IIIT Pet dataset \cite{parkhi2012cats}, a hierarchical classification benchmark.
The dataset contains 7,349 images of cats and dogs, and is divided into 37 fine-grained categories, with approximately 200 images per category.
For each image in the dataset, we first prompt the LVLM to generate detailed image captions.
We then ask the large language model (LLM) to determine whether the generated caption can explicitly identify the corresponding object in the image.
If the LLM can accurately classify images based on these descriptions, it means the image caption effectively captures the detailed visual contents.
Conversely, this implies that information is lost during the image-to-text conversion process via LVLM.
Following ~\cite{costa2007review}, we analyze the classification accuracy of species and breed names in images.

As shown in Table~\ref{tab:problem}, the Species Acc and Breed Acc refer to the accuracy of the LLM in classifying fine-grained and coarse-grained categories based on captions, respectively.
We observe that all models achieve high species accuracy but substantially lower breed accuracy. 
For example, Qwen3-VL-8B~\cite{bai2025qwen2} and InternVL3-8B~\cite{zhu2025internvl3} achieve 100\% species accuracy, with breed accuracy of 40.54\% and 19.29\%, respectively.
This suggests that LVLMs not only tend to describe coarse concepts while ignoring fine details, but also often generate inaccurate descriptions at the finer level, leading to information loss in cross-modal conversion. 
The results indicate that existing LVLMs struggle to generate comprehensive and accurate image captions.

\begin{table}[t]
\tablestyle{0.6pt}{1.0}
    \centering
    \begin{tabular}{lcccc}
        \toprule
        Model & Qwen3-VL-8B~\cite{bai2025qwen2} & InternVL3-8B~\cite{zhu2025internvl3} & LLaVA-v1.6-7B~\cite{liu2024llavanext}\\
        \midrule
      Species Acc   &100.00  &100.00& 99.98 \\
       Breed Acc   &40.54 & 19.29& 15.11\\
        \bottomrule
    \end{tabular}
    \caption{\textbf{Evaluation of fine-grained image classification across LVLMs.} We report species accuracy and breed accuracy on the Oxford-IIIT Pet dataset~\cite{parkhi2012cats}.}
    \label{tab:problem}
\end{table}

\begin{figure*}[t]
    \centering

    \includegraphics[width=1.0\linewidth, trim = 0 0 0 0,clip]{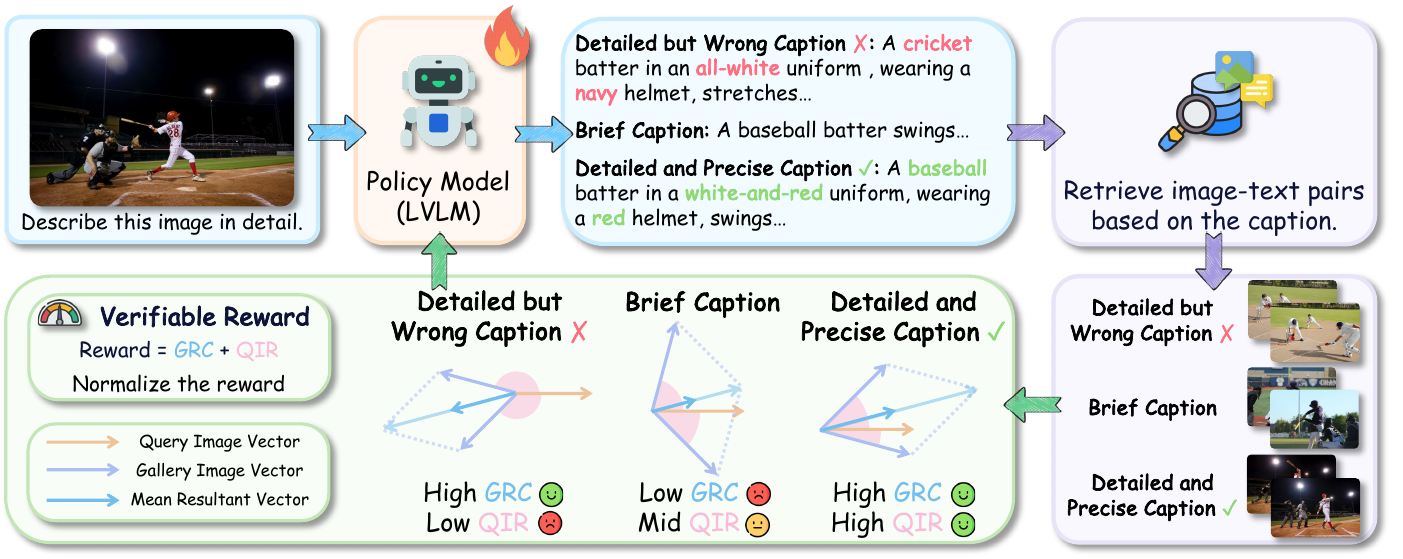}
    \caption{\textbf{The overall architecture of our proposed Cross-modal Identity Mapping (CIM).} We first prompt the LVLM to generate image captions in detail. Based on such captions, CIM then retrieve topK relevant images. In this way, the method analyze the Query-gallery Image Relevance (QIR) and Gallery Representation Consistency (GRC) to infer the information loss between the image and its corresponding caption. Finally, by using CIM as reward function, our method forced LVLMs to generate fine-grained and precise captions.}
    \label{fig:method}
\end{figure*}

\subsection{Quantitative Evaluation of the Information Loss in Modality Conversion}
After verifying the existence of information loss by fine-grained classification labels, we further explore a metric to quantify caption quality without requiring additional human annotations.
We argue that information loss in cross-modal conversion can be quantified by analyzing image similarity in retrieved results with a caption as query.
Intuitively, when we search for images based on a sentence, the more details are included in this sentence, the more consistent the gallery retrieved would be. 
In image captioning, when the captions misrepresent the visual content, the relevance between the source image and the gallery retrieved by that caption is low.

Based on this observation, we present a retrieval-based metric to quantify the information loss in image captioning.
The metric evaluates image caption quality from two dimensions: Gallery Representation Consistency (GRC) and Query-gallery Image Relevance (QIR).
Specifically, $v$ indicates the image and $c$ is the caption generated by the LVLM.
The dataset $\mathcal{D} = \{(x_i, t_i)\}_{i=1}^N$ consists of $N$ samples, each containing an image $x_i$ and its corresponding text $t_i$.
With $c$ as the query, we perform information retrieval on dataset $\mathcal{D}$ via a text retrieval model.
We then select the top K retrieval results as the support set for image caption quantifying, denoted as $A(c) = \operatorname{TopK}\bigl(\{S_i\}_{i=1}^N\bigr) = \{i_1, i_2, \dots, i_K\}$.
$S_{i}$ denotes the cosine similarity between the embedding of the image caption $c$ and the embedding of the $i$-th textual gallery $t_{i}$.
In this way, we use mean resultant length to calculate the Gallery Representation Consistency (GRC) in the support set, defined as follows:
\begin{equation}
GRC(c) = \Bigl\lVert \frac{1}{K}\sum_{r=1}^{K} \tilde v(x_{i_r}) \Bigr\rVert_2,
\end{equation}
The $\tilde v(x_{j})$ refers to the $\ell_2$-normalized embedding extracted by a vision representation model, with the $x_{j}$ as input.
GRC quantifies the internal consistency of images in the support set to indicate the descriptive granularity of image captions.
The low GRC value means the caption is too coarse and cannot retrieve images with similar visual representations.

To quantify the correctness of LVLM-generated captions, we introduce Query-gallery Image Relevance (QIR).
The metric calculates image embedding cosine similarity to analyze images relevance.
Therefore, the relevance between the source image $v$ and the retrieved images $x_{i}$ can be written as $R(v, x_{i})=Cos(\tilde v(v),\tilde v(x_{i}))$.
The formulation of QIR is shown as follows:
\begin{equation}
QIR(v, c) = \sum_{r=1}^{K} \lambda(r) \, R(v, x_{i_r}),
\end{equation}
where $\lambda(r)$ is a exponential decay function.
When the value of QIR is high, which means the retrieved images share common semantic information with the source image $v$, i.e., the query caption accurately describes the visual content.

Moreover, we conduct a verification experiment to further analyze the relationship between our proposed metrics and the information loss in modality conversion.
Specifically, we partition LVLM-generated captions on Oxford–IIIT Pet into multiple bins based on the value of $GRC$ and $QIR$.
For each bin $b$, we calculate the average value of these metrics, denoted $\tilde G_{b}$ and $\tilde Q_{b}$.
Based on the metric mentioned in \cref{sec:3.1}, we can roughly quantify the information loss in image captioning through the logit of the Breed Classification Accuracy $\tilde B_{b}$.
As shown in \cref{fig:correlation}, we regress the correlation between $\tilde Q_{b}$ and $\tilde B_{b}$, as well as between $\tilde G_{b}$ and $\tilde B_{b}$.
The results demonstrate the positive correlation between our proposed metrics and $\tilde B_{b}$.
This correlation verifies that our method quantitatively assesses information loss during modality conversion without additional annotations.

\subsection{Cross-modal Identity Mapping}

Based on the proposed metrics, we introduce a simple yet effective method, Cross-modal Identity Mapping (CIM).
As shown in~\cref{fig:method}, we firstly sample various captions for an input image by LVLMs.
Secondly, the method performs text retrieval using these captions and gets the topk relevant image-text pairs.
By analyzing the relationship among the retrieved images, CIM quantifies the quality of image captions and refines LVLMs using the reward function $\Upsilon (v,c)$.
Specifically, the formulation of our reward function in reinforcement learning is as follows:
\begin{equation}
\begin{aligned}
\Upsilon(v, c)=GRC(c) + \beta\cdot QIR(v, c),
\end{aligned}
\end{equation}
where $\beta$ is a hyperparameter to trade off the balance between precision and detail richness of the image caption.
Following the Group Relative Policy Optimization (GRPO), we sample a group of $G$ captions $\{c_{z}\}_{z=1}^{G}$ from the model, and the advantage $A_{z}$ can be written as:
\begin{equation}
A_{z} = \frac{\Upsilon_{z} - mean(\{\Upsilon_{1},\Upsilon_{2},\dots,\Upsilon_{G}\})}{std(\{\Upsilon_{1},\Upsilon_{2},\dots,\Upsilon_{G}\})}.
\end{equation}

In~\cref{fig:correlation}, we also investigate the correlation between the average value of the reward function $\tilde{R}_{b}$ and the logit of the Breed Accuracy $\tilde{B}_{b}$ in each bin $b$.
Under the supervision of this advantage in reinforcement learning, our method forces LVLMs to generate detailed and accurate image captions.
Accordingly, the information loss between the source image and the LVLM-generated caption is minimized.
In this way, LVLMs strive to establish an identity mapping from visual inputs to textual descriptions during image captioning.

\section{Expriment}
\begin{table*}[!ht] 
\tablestyle{2.0pt}{1.46}
    \centering
    \begin{tabular}{lllcccccccc}
    \hline
       \multirow{2}{*}{\makecell{Base \\Model}} &  
       \multirow{2}{*}{Bench} &  
       \multirow{2}{*}{Method} & 
        \multirow{2}{*}{CAPTURE} & 
       \multicolumn{3}{c}{Objects} & 
       \multicolumn{3}{c}{Attributes} & 
       Relations  \\ 
       \cline{5-11}
       &&&& Precision & Recall & F1 & Precision & Recall & F1 & QA  \\ \hline
       
       \multirow{4}{*}{\makecell{LLaVA1.5\\-7B\\~\cite{liu2023improvedllava}}} & 
       \multirow{2}{*}{\makecell{COCO \\-LN500}} & 
       \vb Base & 44.75 & 81.20 & 59.02 & 67.56 & 58.61 & 37.2 & 42.34 & 14.38 \\ 
       && \oursbg \cb Ours & \oursbg 48.62 ($\uparrow$3.9) & \oursbg 83.84 ($\uparrow$2.6) & \oursbg 60.70 ($\uparrow$1.7) & \oursbg 69.80 ($\uparrow$2.2) & \oursbg 71.24 ($\uparrow$12.6) & \oursbg 48.91 ($\uparrow$11.7) & \oursbg 54.38 ($\uparrow$12.0) & \oursbg 24.98 ($\uparrow$10.6) \\ 

       & 
       \multirow{2}{*}{\makecell{DOCCI \\500}} & 
       \vb Base & 50.72 & 81.13 & 48.37 & 59.41 & 64.12 & 39.78 & 48.01 & 9.19 \\ 
       && \oursbg \cb Ours & \oursbg 57.10 ($\uparrow$6.4) & \oursbg 83.12 ($\uparrow$2.0) & \oursbg 52.78 ($\uparrow$4.4) & \oursbg 63.38 ($\uparrow$4.0) & \oursbg 72.06 ($\uparrow$7.9) & \oursbg 47.42 ($\uparrow$7.6) & \oursbg 56.28 ($\uparrow$8.3) & \oursbg 19.87 ($\uparrow$10.7) \\ \hline

       \multirow{4}{*}{\makecell{Qwen2\\-VL-7B\\~\cite{wang2024qwen2}}} & 
       \multirow{2}{*}{\makecell{COCO \\-LN500}} & 
       \vb Base & 46.52 & 81.12 & 61.82 & 69.47 & 66.48 & 42.86 & 48.68 & 20.47 \\
       && \oursbg \cb Ours & \oursbg 48.64 ($\uparrow$2.1) & \oursbg 80.90 ($\downarrow$ 0.2) & \oursbg 72.23 ($\uparrow$10.4) & \oursbg 75.80 ($\uparrow$6.3) & \oursbg 72.45 ($\uparrow$6.0) & \oursbg 54.08 ($\uparrow$11.2) & \oursbg 58.22 ($\uparrow$9.5) & \oursbg 38.71 ($\uparrow$18.2) \\ 

       & 
       \multirow{2}{*}{\makecell{DOCCI \\500}} & 
       \vb Base & 57.96 & 83.69 & 56.79 & 66.47 & 69.96 & 43.27 & 52.65 & 17.57 \\
       && \oursbg \cb Ours & \oursbg 63.12 ($\uparrow$5.2) & \oursbg 81.99 ($\downarrow$ 1.7) & \oursbg 64.61 ($\uparrow$7.8) & \oursbg 71.43 ($\uparrow$5.0) & \oursbg 74.49 ($\uparrow$4.5) & \oursbg 50.19 ($\uparrow$6.9) & \oursbg 59.18 ($\uparrow$6.5) & \oursbg 32.12 ($\uparrow$14.5) \\ \hline

       \multirow{4}{*}{\makecell{Qwen2.5\\-VL-7B\\~\cite{bai2025qwen2}}} & 
       \multirow{2}{*}{\makecell{COCO \\-LN500}} & 
       \vb Base & 44.12 & 82.35 & 55.72 & 65.37 & 66.30 & 39.90 & 46.25 & 23.76 \\ 
       && \oursbg \cb Ours & \oursbg 48.93 ($\uparrow$4.8) & \oursbg 80.31 ($\downarrow$ 2.0) & \oursbg 75.91 ($\uparrow$20.2) & \oursbg 77.59 ($\uparrow$12.2) & \oursbg 72.49 ($\uparrow$6.2) & \oursbg 54.46 ($\uparrow$14.6) & \oursbg 58.51 ($\uparrow$12.3) & \oursbg 44.15 ($\uparrow$20.4) \\ 

       & 
       \multirow{2}{*}{\makecell{DOCCI \\500}} & 
       \vb Base & 55.89 & 84.64 & 54.96 & 65.06 & 72.15 & 42.13 & 52.27 & 24.35 \\ 
       && \oursbg \cb Ours & \oursbg 63.46 ($\uparrow$7.6) & \oursbg 79.28 ($\downarrow$ 5.4) & \oursbg 66.97 ($\uparrow$12.0) & \oursbg 71.88 ($\uparrow$6.8) & \oursbg 73.52 ($\uparrow$1.4) & \oursbg 50.45 ($\uparrow$8.3) & \oursbg 59.08 ($\uparrow$6.8) & \oursbg 34.70 ($\uparrow$10.4) \\ \hline

       \multirow{4}{*}{\makecell{InternVL2\\-8B\\~\cite{chen2024internvl}}} & 
       \multirow{2}{*}{\makecell{COCO \\-LN500}} & 
       \vb Base & 45.86 & 80.46 & 65.20 & 71.05 & 70.63 & 47.40 & 52.73 & 26.65 \\ 
       && \oursbg \cb Ours & \oursbg 49.06 ($\uparrow$3.2) & \oursbg 82.05 ($\uparrow$1.6) & \oursbg 69.40 ($\uparrow$4.2) & \oursbg 74.57 ($\uparrow$3.5) & \oursbg 73.50 ($\uparrow$2.9) & \oursbg 52.59 ($\uparrow$5.2) & \oursbg 57.49 ($\uparrow$4.8) & \oursbg 35.54 ($\uparrow$8.9) \\ 

       & 
       \multirow{2}{*}{\makecell{DOCCI \\500}} & 
       \vb Base & 58.83 & 81.49 & 59.54 & 67.72 & 70.84 & 44.24 & 53.51 & 22.65 \\ 
       && \oursbg \cb Ours & \oursbg 60.82 ($\uparrow$2.0) & \oursbg 82.16 ($\uparrow$0.7) & \oursbg 60.35 ($\uparrow$0.8) & \oursbg 68.66 ($\uparrow$0.9) & \oursbg 74.16 ($\uparrow$3.3) & \oursbg 47.51 ($\uparrow$3.3) & \oursbg 56.98 ($\uparrow$3.5) & \oursbg 26.36 ($\uparrow$3.7) \\ \hline

       \multirow{4}{*}{\makecell{InternVL2.5\\-8B\\~\cite{chen2024expanding,wang2024mpo}}} & 
       \multirow{2}{*}{\makecell{COCO \\-LN500}} & 
       \vb Base & 47.09 & 82.42 & 64.02 & 71.42 & 72.00 & 46.03 & 52.50 & 27.29 \\ 
       && \oursbg \cb Ours & \oursbg 48.28 ($\uparrow$1.2) & \oursbg 80.31 ($\downarrow$ 2.1) & \oursbg 70.42 ($\uparrow$6.4) & \oursbg 74.53 ($\uparrow$3.1) & \oursbg 71.85 ($\downarrow$ 0.2) & \oursbg 51.71 ($\uparrow$5.7) & \oursbg 56.07 ($\uparrow$3.6) & \oursbg 36.96 ($\uparrow$9.7) \\ 

       & \multirow{2}{*}{\makecell{DOCCI \\500}} & 
       \vb Base & 58.64 & 84.11 & 58.07 & 67.62 & 73.48 & 44.79 & 54.69 & 24.63 \\ 
       && \oursbg \cb Ours & \oursbg 60.75 ($\uparrow$2.1) & \oursbg 82.17 ($\downarrow$ 1.9) & \oursbg 62.28 ($\uparrow$4.2) & \oursbg 69.99 ($\uparrow$2.4) & \oursbg 72.96 ($\downarrow$ 0.5) & \oursbg 46.64 ($\uparrow$1.9) & \oursbg 56.07 ($\uparrow$1.4) & \oursbg 28.90 ($\uparrow$4.3) \\ \hline

       \multirow{4}{*}{\makecell{InternVL3\\-8B\\~\cite{zhu2025internvl3}}} & 
       \multirow{2}{*}{\makecell{COCO \\-LN500}} & 
       \vb Base & 47.88 & 82.51 & 63.31 & 71.00 & 71.14 & 43.97 & 50.66 & 26.44 \\ 
       && \oursbg \cb Ours & \oursbg 48.90 ($\uparrow$1.0) & \oursbg 80.33 ($\downarrow$ 2.2) & \oursbg 73.33 ($\uparrow$10.0) & \oursbg 76.14 ($\uparrow$5.1) & \oursbg 73.33 ($\uparrow$2.2) & \oursbg 54.54 ($\uparrow$10.6) & \oursbg 58.70 ($\uparrow$8.0) & \oursbg 38.67 ($\uparrow$12.2) \\ 

       & 
       \multirow{2}{*}{\makecell{DOCCI \\500}} & 
       \vb Base & 56.95 & 85.98 & 55.39 & 66.08 & 74.75 & 43.19 & 53.72 & 25.11 \\ 
       && \oursbg \cb Ours & \oursbg 62.18 ($\uparrow$5.2) & \oursbg 82.73 ($\downarrow$ 3.2) & \oursbg 62.86 ($\uparrow$7.5) & \oursbg 70.47 ($\uparrow$4.4) & \oursbg 76.22 ($\uparrow$1.5) & \oursbg 49.58 ($\uparrow$6.4) & \oursbg 59.26 ($\uparrow$5.5) & \oursbg 30.39 ($\uparrow$5.3) \\ \hline 
    \end{tabular}

    \caption{
    \textbf{  Performance of Reinforcement Learning with CIM on the Base Model 
 over COCO-LN500~\cite{pont2020connecting} and DOCCI500~\cite{onoe2024docci}. } 
 $\uparrow$/$\downarrow$ indicate improvement/degradation over \textbf{the Base Model} (higher is better). 
    The results indicate that post-training the base model with our CIM enhances its ability to generate high-quality captions. 
  }
        \label{Exp:table_not_sft}
\end{table*}

In this section, we first describe our experimental settings. 
Subsequently, we present the model’s performance before
and after the application of our Cross-modal Identity Mapping (CIM) across various Large Vision-Language Models (LVLMs). 
Following this, we conduct an ablation study to verify the effectiveness of each metric we proposed.
Moreover, we conduct the data scale-up experiment to investigate the scalability of the retrieval corpus of our method.
Finally, we evaluate model performance based on different retrieval models to demonstrate the robustness of our method.
Additional results about the performance of our method can be found in the supplementary materials. 

\subsection{Experimental Settings}
\textbf{Training settings.}
Following SC-Captioner~\cite{zhang2025sc}, we perform reinforcement learning on images in the RefinedCaps dataset~\cite{zhang2025sc}, which comprises 6.5K images sampled from the COCO training split~\cite{lin2014microsoft}. 
For each image, we prompt LVLMs to generate $G=5$ distinct captions. 
For text retrieval, we employ SBERT text encoder~\cite{reimers2019sentence} with MPNet-base backbone~\cite{song2020mpnet}. 
We retrieve the top-K candidates with $K=5$ from a dataset constructed by augmenting RefinedCaps~\cite{zhang2025sc} with DenseFusion-1M~\cite{li2024densefusion}. 
For images, we use OpenCLIP ViT-H/14~\cite{cherti2023reproducible} as the image encoder.
When computing the $QIR$ metric, we set the exponential decay function to $\lambda(r)=1/2^{r-1}$. In the reward function, we set the trade-off parameter to $\beta=1$.
We adopt the VERL framework~\cite{sheng2025hybridflow} for training. 
During the training phase, we use the Adam optimizer and train for two epochs with an initial learning rate of $1\times10^{-6}$, a batch size of 256, and a mini-batch size of 64. 

\noindent \textbf{Models.}
We conduct experiments across multiple LVLMs to demonstrate the generalizability of our proposed CIM. 
Specifically, we evaluate on three architecturally distinct models: LLaVA~\cite{liu2023improvedllava,liu2024llavanext}, Qwen-VL~\cite{bai2025qwen2,wang2024qwen2}, and InternVL~\cite{chen2024internvl,chen2024expanding,wang2024mpo,zhu2025internvl3}. Furthermore, we also compare different model versions and scales. 

\noindent \textbf{Benchmarks.}
We select DOCCI500~\cite{onoe2024docci} and COCO-LN500~\cite{pont2020connecting} as test datasets
to effectively evaluate image captioning performance of LVLMs. 
DOCCI500~\cite{onoe2024docci} is a random sample of 500 image–caption pairs from the DOCCI~\cite{onoe2024docci} test split and is largely non-human-centric. 
To complement DOCCI500, COCO-LN500~\cite{pont2020connecting} consists of 500 images from the Localized-narratives~\cite{pont2020connecting} test split in COCO2017~\cite{lin2014microsoft}, filtered for human-related descriptions.

\noindent \textbf{Metrics. }
We evaluate overall caption quality using CAPTURE~\cite{dong2024benchmarking}, which parses ground-truth and generated captions into objects, attributes, and relations, computes an F1 score for each component, and aggregates them via a weighted sum. 
Moreover, to further assess the precision and detail of the captions, we evaluate fine-grained metrics for objects, attributes, and relations, similar to SC-Captioner~\cite{zhang2025sc}.  
The large language model (LLM) used in relation evaluation and \cref{sec:3.1} is Qwen3~\cite{yang2025qwen3}.

\noindent \textbf{Baselines. }
We consider two reinforcement learning settings to validate the effectiveness of our proposed CIM: reinforcement learning on the base model and reinforcement learning on the SFT model.
In the first setting, CIM is applied directly to the pre-trained LVLM.
In the second setting, CIM is applied to the SFT model.

\begin{table*}[!htbp]
\tablestyle{2.5pt}{1.3}
    \centering
    \begin{tabular}{lllcccccccc}
    \hline
       \multirow{2}{*}{\makecell{Base \\Model}} &  
       \multirow{2}{*}{\makecell{Bench}} &  
       \multirow{2}{*}{Method} & 
        \multirow{2}{*}{CAPTURE} & 
       \multicolumn{3}{c}{Objects} & 
       \multicolumn{3}{c}{Attributes} & 
       Relations  \\ 
       \cline{5-11}
       &&&& Precision & Recall & F1 & Precision & Recall & F1 & QA  \\ \hline
       
       \multirow{6}{*}{\makecell{LLaVA1.5\\-7B\\~\cite{liu2023improvedllava}}} & 
       \multirow{3}{*}{\makecell{COCO \\-LN500}}
       & 
       \vb SFT & 46.62&77.09 & 71.16 & 73.45 & 66.65 & 50.57 & 54.25 & 28.59 \\
        && \vb SC-Captioner~\cite{zhang2025sc} &47.11& 76.43 & 75.65 & 75.20 & 65.79 & 53.01 & 55.35 & 33.63 \\
       && \oursbg \cb SFT + Ours  & \oursbg47.74($\uparrow$0.6)&
          \oursbg 79.48 ($\uparrow$3.1) & 
          \oursbg 69.75 ($\downarrow$5.9) & 
          \oursbg 73.77 ($\downarrow$1.4) & 
          \oursbg 70.48 ($\uparrow$4.7) & 
          \oursbg 51.29 ($\downarrow$1.7) & 
          \oursbg 56.04 ($\uparrow$0.7) & 
          \oursbg 33.96 ($\uparrow$0.3) \\
          \cline{2-11}

       & \multirow{3}{*}{\makecell{DOCCI \\500}} & 
    \vb    SFT& 60.92& 78.15 & 62.20 & 68.44 & 66.85 & 46.27 & 53.93 & 19.87 \\
        && \vb SC-Captioner~\cite{zhang2025sc} &62.29& 77.58 & 66.05 & 70.30 & 67.64 & 50.70 & 57.10 & 22.69 \\
       && \oursbg  \cb SFT + Ours & \oursbg62.39($\uparrow$0.1) &
          \oursbg 80.28 ($\uparrow$2.7) & 
          \oursbg 64.09 ($\downarrow$2.0) & 
          \oursbg 70.55 ($\uparrow$0.3) & 
          \oursbg 71.31 ($\uparrow$3.7) & 
          \oursbg 47.86 ($\downarrow$2.8) & 
          \oursbg 56.48 ($\downarrow$0.6) & 
          \oursbg 25.88 ($\uparrow$3.2) \\ \hline

       \multirow{6}{*}{\makecell{Qwen2 \\-VL-7B\\~\cite{wang2024qwen2}}} & 
       \multirow{3}{*}{\makecell{COCO \\-LN500}} & 
    \vb    SFT & 47.14&78.64 & 73.24 & 75.37 & 69.01 & 53.15 & 56.54 & 36.39 \\
      &&   \vb SC-Captioner~\cite{zhang2025sc} & 47.51&78.72 & 75.01 & 76.37 & 68.43 & 55.11 & 57.56 & 38.51 \\
     && \oursbg  \cb  SFT + Ours  & \oursbg48.61($\uparrow$1.1) &
          \oursbg 79.14 ($\uparrow$0.4) & 
          \oursbg 75.15 ($\uparrow$0.1) & 
          \oursbg 76.65 ($\uparrow$0.3) & 
          \oursbg 70.72 ($\uparrow$2.3) & 
          \oursbg 54.38 ($\downarrow$0.7) & 
          \oursbg 58.09 ($\uparrow$0.5) & 
          \oursbg 42.12 ($\uparrow$3.6) \\
          \cline{2-11}

       & \multirow{3}{*}{\makecell{DOCCI \\500}} & 
    \vb    SFT &62.05& 79.99 & 62.68 & 69.50 & 68.85 & 47.57 & 55.50 & 27.65 \\
        && \vb SC-Captioner~\cite{zhang2025sc} &63.34
        &80.20 & 66.00 & 71.63 & 69.54 & 50.34 & 57.67 & 30.51 \\
     && \oursbg  \cb  SFT + Ours  & \oursbg64.31($\uparrow$1.0) &
          \oursbg 80.16 ($\downarrow$0.0) & 
          \oursbg 69.58 ($\uparrow$3.6) & 
          \oursbg 73.87 ($\uparrow$2.2) & 
          \oursbg 72.69 ($\uparrow$3.2) & 
          \oursbg 50.36 ($\uparrow$0.0) & 
          \oursbg 58.68 ($\uparrow$1.0) & 
          \oursbg 36.32 ($\uparrow$5.8) \\  \hline
    \end{tabular}
    \caption{\textbf{Performance of Reinforcement Learning with CIM on the SFT Model over COCO-LN500~\cite{pont2020connecting} and DOCCI500~\cite{onoe2024docci}.} 
    $\uparrow$/$\downarrow$ indicate improvement/degradation over \textbf{SC-Captioner}~\cite{zhang2025sc} (higher is better). 
    Our method consistently improves relation understanding and attribute precision, with minor trade-offs in object recall.}
    \label{Exp:table_sft}
\end{table*}

\subsection{Reinforcement Learning on the Base Model }
Cross-modal Identity Mapping (CIM) can boost the precision and detail richness of captions generated by LVLMs via reinforcement learning.
To this end, we conduct experiments to optimize LVLMs and test the model performance on benchmarks to verify the effectiveness of our method.

As shown in \cref{Exp:table_not_sft}, 
Base refers to the base LVLMs, without any post-training or further adaptation. 
`Ours' denotes the base LVLMs trained solely with our proposed CIM via reinforcement learning, without any intermediate supervised fine-tuning. 
The results show that our approach yields consistent gains across all evaluation settings. 
When evaluating the model on more challenging tasks: Attributes and Relations, our method outperforms the base model across all metrics by a large margin. 
For instance, when performing reinforcement learning on Qwen2.5-VL-7B~\cite{bai2025qwen2}, our method achieves improvements by 12.3\% in the F1 score on Attributes testing and 20.4\% in the QA score on Relations evaluation. 
However, the results show that our method has lower precision compared with the base model when tested on Objects on QwenVL~\cite{bai2025qwen2,wang2024qwen2} and InternVL~\cite{chen2024internvl,chen2024expanding,wang2024mpo,zhu2025internvl3}. 
These results indicate that the GRPO improves caption quality based on a global-level reward score but fails to capture nuances in subtasks such as object recognition.
During reinforcement learning, LVLM tends to improve model performance on more challenging tasks (e.g., Attributes and Relations) to obtain a more significant bonus.
When testing models with the CAPTURE metric, we achieve an average improvement of 3.7\% over the base model across two benchmarks. 
Moreover, our method can even outperform that of the 72B model (see the supplementary materials for details).
The above results demonstrate that our proposed Cross-modal Identity Mapping enhances the model’s capacity to capture diverse fine-grained visual cues and integrate them into the image captioning.

\subsection{Reinforcement Learning on the SFT Model }
We conduct additional experiments to demonstrate the effectiveness of our method on strong baselines.
To this end, we investigate the performance of LVLMs fine-tuned on image captioning datasets.
Moreover, we perform reinforcement learning on these LVLMs using our method, Cross-modal Identity Mapping (CIM).

As shown in \cref{Exp:table_sft}, SFT denotes the LVLMs supervised fine-tuned on the RefinedCaps~\cite{zhang2025sc} dataset. 
SC-Captioner~\cite{zhang2025sc} applies reinforcement learning on the SFT models.
SFT+Ours represents the SFT models post-trained with our proposed CIM. 
The experimental results show that even the SFT models achieve significant improvements compared with the Base models reported in \cref{Exp:table_not_sft}, our method can further enhance their performance during the post-training phase. 
Specifically, on the Relations metric of two popular benchmarks, our method achieves an average improvement of 6.4\% over SFT and 3.2\% over SC-Captioner~\cite{zhang2025sc}.
In particular, our method achieved more significant improvements when using the more powerful LVLM, i.e., Qwen2-VL-7B~\cite{bai2025qwen2}. 
This is especially evident in the CAPTURE metric, where our method outperforms SC-Captioner~\cite{zhang2025sc} by 1.1\% and 1.0\% on COCO-LN500~\cite{pont2020connecting} and DOCCI500~\cite{onoe2024docci}, respectively. 
In contrast, when applied to LLaVA1.5-7B~\cite{liu2023improvedllava}, the average improvement on CAPTURE is only 0.4\%. 
This trend can be attributed to the nature of GRPO~\cite{shao2024deepseekmath}, which relies on multiple rollouts. Stronger models are more capable of efficiently sampling high-quality outputs, thereby benefiting more from the reinforcement learning process.
These results further validate that the stronger the base model, the more significant its improvement during the post-training~\cite{zhang2025will} phase. 

\begin{figure}[t]
    \centering
    \includegraphics[width=\linewidth]{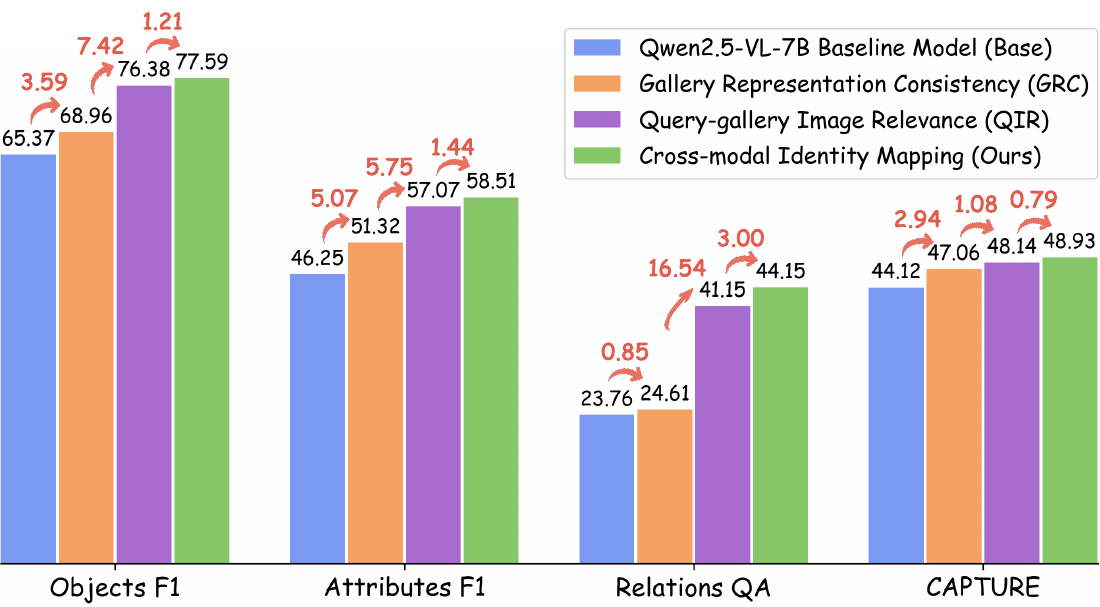}
    \caption{\textbf{Ablation study of reward components on COCO-LN500~\cite{pont2020connecting} using Qwen2.5-VL-7B~\cite{bai2025qwen2}. }}
    \label{fig:ablation}
\end{figure}

Moreover, our method achieves significantly larger improvements on the challenging sub-tasks of Attributes and Relations by 0.4\% and 3.2\% on average compared to SC-Captioner~\cite{zhang2025sc}. 
In contrast, it gets lower precision on Objects, improving by only 0.3\%, a task already handled well by the SFT model. 
This behavior is consistent with the findings in \cref{Exp:table_not_sft} and further supports the above argument.
Specifically, LVLMs focus on tasks (i.e., Relations and Attributes) with greater room for improvement to achieve higher reward scores during the reinforcement learning phase.
In summary, the results indicate the superiority of our method, even when training on strong baselines.

\subsection{Ablation Study}
To validate the effectiveness of each our proposed metrics, we design dedicated ablation studies and provide thorough analysis in this part. 
As shown in ~\cref{fig:ablation}, Base is the base model without any post-training. 
QIR and GRC refer to LVLMs optimized by reinforcement learning that use the Query-gallery Image Relevance (QIR) and Gallery Representation Consistency (GRC) as their reward function.
Ours means that LVLMs are post-trained based on Cross-modal Identity Mapping (CIR), combined with QIR and GRC. 

It can be observed that each reward method achieves consistent improvements in image captioning compared with the base model. 
For example, on COCO-LN500~\cite{pont2020connecting}, LVLMs post-trained with GRC achieved improvements of 2.9\%, 3.6\%, 5.1\%, and 0.9\% on CAPTURE, Objects F1, Attributes F1, and Relations QA, respectively.
Moreover, QIR achieves improvements of 4.0\%, 11.0\%, 10.8\%, and 17.4\%, respectively.
Such results demonstrate that GRC and QIR quantitatively evaluates the precision and detail richness of captions.
Most importantly, when combining QIR and GRC, our method achieves the best performance across various metrics.
This demonstrates that QIR and GRC are complementary: jointly optimizing them leads to more detailed and precise caption.

\subsection{Scalability of the Retrieval Corpus}
\begin{figure}[t]
    \centering
    \includegraphics[width=\linewidth]{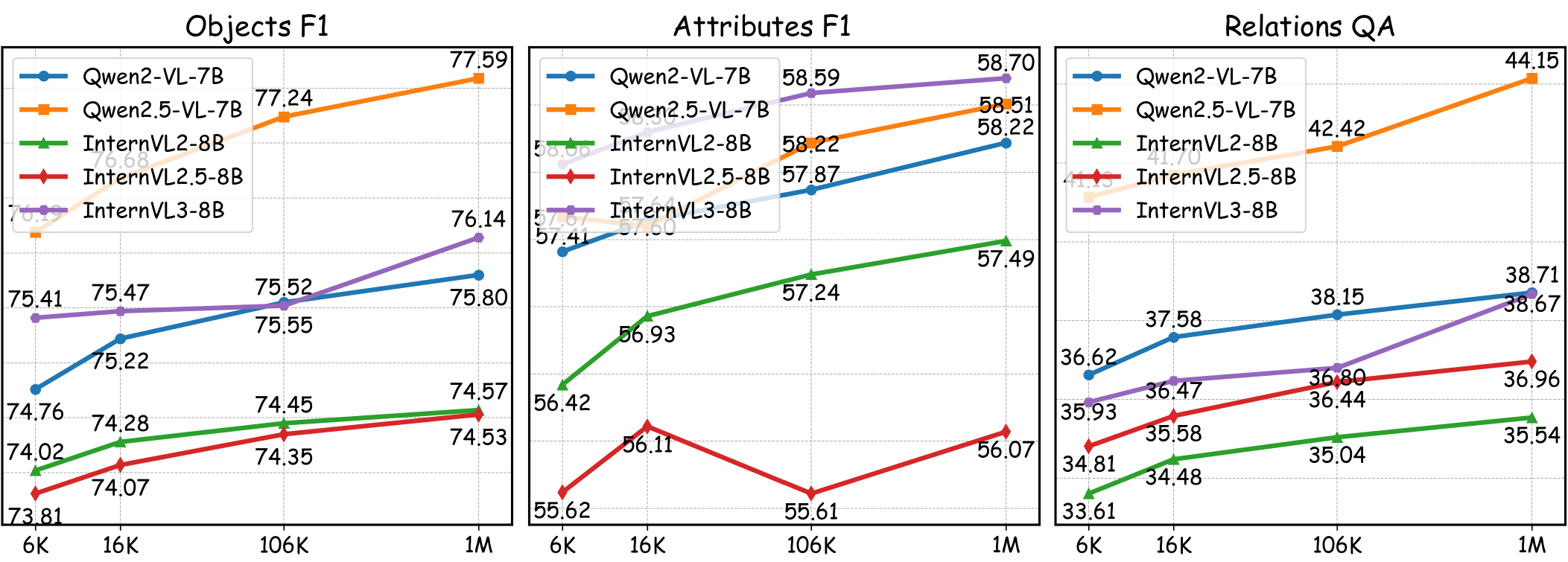}
    \caption{\textbf{Scalability study of the retrieval corpus on COCO-LN500~\cite{pont2020connecting} across multiple LVLMs.}}
    \label{fig:scalability}
\end{figure}

We argue that as the retrieval corpus grows, the number of hard negative samples for each image increases accordingly. 
This makes it necessary to use more detailed and precise captions to retrieve consistent and relevant images. 
To verify the scalability of the retrieval corpus, we train on retrieval corpora of different sizes and evaluate on COCO-LN500~\cite{pont2020connecting} across multiple LVLMs. Specifically, we consider RefinedCaps~\cite{zhang2025sc}, RefinedCaps augmented with DenseFusion-4V-10K randomly selected from DenseFusion-4V-100K~\cite{li2024densefusion}, RefinedCaps augmented with DenseFusion-4V-100K~\cite{li2024densefusion}, and RefinedCaps augmented with DenseFusion-1M~\cite{li2024densefusion}.
As shown in \cref{fig:scalability}, performance improves as the corpus size grows and exhibits potential for further improvement. 
Specifically, on Qwen2.5-VL-7B~\cite{bai2025qwen2}, scaling from RefinedCaps to RefinedCaps augmented with DenseFusion-4V-100K brings improvements of 1.1\%, 0.6\%, and 1.3\% on Objects F1, Attributes F1, and Relations QA, respectively. 
Scaling further to RefinedCaps augmented with DenseFusion-1M brings additional improvements of 0.4\%, 0.3\%, and 1.7\%. 
These results demonstrate that as the retrieval corpus grows, performance on Objects F1, Attributes F1, and Relations QA improves, confirming that LVLMs must generate more detailed and precise captions to retrieve consistent and relevant images. 

\begin{table}[t]
\tablestyle{1.5pt}{1.4}
\centering
\begin{tabular}{llcccc}
\hline
\multirow{2}{*}{\makecell{Image \\Encoder}} & 
\multirow{2}{*}{\makecell{Text \\Encoder}} & 

\multirow{2}{*}{CAPTURE} & 
Objects & 
Attributes & 
Relations \\ \cline{4-6}
&&& F1 & F1 & QA \\\hline
\multirow{2}{*}{\makecell{DINOv3 \\ViT-L/16~\cite{simeoni2025dinov3}}}
&  MiniLM~\cite{wang2020minilm} & 48.17 & 76.62 & 59.05 & 41.03 \\
& \oursbg MPNet~\cite{song2020mpnet} & \oursbg 48.64 & \oursbg 76.32 & \oursbg 59.22 & \oursbg 42.49 \\
\multirow{2}{*}{\makecell{OpenCLIP \\ViT-H/14~\cite{cherti2023reproducible}}} 
&  MiniLM~\cite{wang2020minilm} & 48.99 & 76.91 &   58.96 & 41.16  \\
& \oursbg MPNet~\cite{song2020mpnet} & \oursbg  48.93 & \oursbg  77.59 & \oursbg  58.51 & \oursbg 44.15 \\\hline
\end{tabular}
\caption{\textbf{Robustness study of diverse retrieval models on COCO-LN500~\cite{pont2020connecting} using Qwen2.5-VL-7B~\cite{bai2025qwen2}.}}
\label{tab:robustness}
\end{table}

\subsection{Robustness across Diverse Retrieval Models}
To evaluate the impact of different encoders in the retrieval process on our proposed method, we evaluate performance under various combinations of image and text encoders. 
As shown in \cref{tab:robustness}, we assess CIM across four popular representative encoders: DINOv3 ViT-L/16~\cite{simeoni2025dinov3} or OpenCLIP ViT-H/14 ~\cite{cherti2023reproducible} as the image encoder, and SBERT~\cite{reimers2019sentence} with either a MiniLM-base~\cite{wang2020minilm} or MPNet-base~\cite{song2020mpnet} backbone as the text encoder. 
The evaluation covers multiple subtasks, including CAPTURE, Objects, Attributes, and Relations. 
The results reveal minimal performance variance across configurations. 
For instance, CAPTURE, Objects F1, and Attributes F1 scores vary by no more than 0.8\%, 1.2\%, and 0.7\%, respectively, on COCO-LN500~\cite{pont2020connecting}, indicating statistically insignificant differences. 
This demonstrates that CIM is highly stable with respect to potential information loss introduced by different encoding schemes, and its quantification is largely insensitive to the choice of underlying encoders. 
Consequently, CIM exhibits strong encoder-agnostic robustness, enabling practical deployment without the need for adjustment when swapping encoders. 
\section{Conclusion}
\label{sec:conclusion}
In this paper, using the caption as a query, we demonstrated that retrieved images serve as indicators of information loss between the source image and its corresponding caption.
Based on this novel insight, we quantitatively evaluate the quality of image captions without requiring additional annotations.
In this way, we further proposed Cross-modal Identity Mapping (CIM), a reinforcement learning algorithm that boosts the precision and detail richness of LVLM-generated captions by minimizing information loss in modality conversion.
Extensive experiments demonstrate that our proposed method forced LVLMs to establish an identity mapping from visual inputs to textual outputs, achieving superior performance in image captioning.
In summary, this research provides a new perspective for improving image captioning performance of LVLMs, shedding new light on the development of this field. 
{
    \small
    \bibliographystyle{ieeenat_fullname}
    \bibliography{main}

\begin{thebibliography}{86}
\providecommand{\natexlab}[1]{#1}
\providecommand{\url}[1]{\texttt{#1}}
\expandafter\ifx\csname urlstyle\endcsname\relax
  \providecommand{\doi}[1]{doi: #1}\else
  \providecommand{\doi}{doi: \begingroup \urlstyle{rm}\Url}\fi

\bibitem[An et~al.(2024)An, Tian, Leng, Nie, Lin, Wang, Dai, Chen, and Lu]{an2024agla}
Wenbin An, Feng Tian, Sicong Leng, Jiahao Nie, Haonan Lin, QianYing Wang, Guang Dai, Ping Chen, and Shijian Lu.
\newblock Agla: Mitigating object hallucinations in large vision-language models with assembly of global and local attention.
\newblock \emph{arXiv preprint arXiv:2406.12718}, 2024.

\bibitem[Bahng et~al.(2025)Bahng, Chan, Durand, and Isola]{bahng2025cycle}
Hyojin Bahng, Caroline Chan, Fredo Durand, and Phillip Isola.
\newblock Cycle consistency as reward: Learning image-text alignment without human preferences.
\newblock \emph{arXiv preprint arXiv:2506.02095}, 2025.

\bibitem[Bai et~al.(2023)Bai, Bai, Yang, Wang, Tan, Wang, Lin, Zhou, and Zhou]{bai2023qwen}
Jinze Bai, Shuai Bai, Shusheng Yang, Shijie Wang, Sinan Tan, Peng Wang, Junyang Lin, Chang Zhou, and Jingren Zhou.
\newblock Qwen-vl: A frontier large vision-language model with versatile abilities.
\newblock \emph{arXiv preprint arXiv:2308.12966}, 2023.

\bibitem[Bai et~al.(2025)Bai, Chen, Liu, Wang, Ge, Song, Dang, Wang, Wang, Tang, et~al.]{bai2025qwen2}
Shuai Bai, Keqin Chen, Xuejing Liu, Jialin Wang, Wenbin Ge, Sibo Song, Kai Dang, Peng Wang, Shijie Wang, Jun Tang, et~al.
\newblock Qwen2. 5-vl technical report.
\newblock \emph{arXiv preprint arXiv:2502.13923}, 2025.

\bibitem[Chen et~al.(2024{\natexlab{a}})Chen, Li, Dong, Zhang, He, Wang, Zhao, and Lin]{chen2024sharegpt4v}
Lin Chen, Jinsong Li, Xiaoyi Dong, Pan Zhang, Conghui He, Jiaqi Wang, Feng Zhao, and Dahua Lin.
\newblock Sharegpt4v: Improving large multi-modal models with better captions.
\newblock In \emph{European Conference on Computer Vision}, pages 370--387. Springer, 2024{\natexlab{a}}.

\bibitem[Chen et~al.(2024{\natexlab{b}})Chen, Ma, Zhang, Qi, Yuan, Li, Pu, Shan, Qi, and Hu]{chen2024make}
Yuxin Chen, Zongyang Ma, Ziqi Zhang, Zhongang Qi, Chunfeng Yuan, Bing Li, Junfu Pu, Ying Shan, Xiaojuan Qi, and Weiming Hu.
\newblock How to make cross encoder a good teacher for efficient image-text retrieval?
\newblock In \emph{Proceedings of the IEEE/CVF Conference on Computer Vision and Pattern Recognition}, pages 26994--27003, 2024{\natexlab{b}}.

\bibitem[Chen et~al.(2024{\natexlab{c}})Chen, Wang, Cao, Liu, Gao, Cui, Zhu, Ye, Tian, Liu, et~al.]{chen2024expanding}
Zhe Chen, Weiyun Wang, Yue Cao, Yangzhou Liu, Zhangwei Gao, Erfei Cui, Jinguo Zhu, Shenglong Ye, Hao Tian, Zhaoyang Liu, et~al.
\newblock Expanding performance boundaries of open-source multimodal models with model, data, and test-time scaling.
\newblock \emph{arXiv preprint arXiv:2412.05271}, 2024{\natexlab{c}}.

\bibitem[Chen et~al.(2024{\natexlab{d}})Chen, Wu, Wang, Su, Chen, Xing, Zhong, Zhang, Zhu, Lu, et~al.]{chen2024internvl}
Zhe Chen, Jiannan Wu, Wenhai Wang, Weijie Su, Guo Chen, Sen Xing, Muyan Zhong, Qinglong Zhang, Xizhou Zhu, Lewei Lu, et~al.
\newblock Internvl: Scaling up vision foundation models and aligning for generic visual-linguistic tasks.
\newblock In \emph{Proceedings of the IEEE/CVF Conference on Computer Vision and Pattern Recognition}, pages 24185--24198, 2024{\natexlab{d}}.

\bibitem[Chen et~al.(2024{\natexlab{e}})Chen, Zhao, Luo, Yao, Li, and Zhou]{chen2024halc}
Zhaorun Chen, Zhuokai Zhao, Hongyin Luo, Huaxiu Yao, Bo Li, and Jiawei Zhou.
\newblock Halc: Object hallucination reduction via adaptive focal-contrast decoding.
\newblock \emph{arXiv preprint arXiv:2403.00425}, 2024{\natexlab{e}}.

\bibitem[Cheng et~al.(2025)Cheng, Zhang, Zhang, Yang, Guan, Wu, Li, Zhang, Liu, Mai, et~al.]{cheng2025simplevqa}
Xianfu Cheng, Wei Zhang, Shiwei Zhang, Jian Yang, Xiangyuan Guan, Xianjie Wu, Xiang Li, Ge Zhang, Jiaheng Liu, Yuying Mai, et~al.
\newblock Simplevqa: Multimodal factuality evaluation for multimodal large language models.
\newblock In \emph{Proceedings of the IEEE/CVF International Conference on Computer Vision}, pages 4637--4646, 2025.

\bibitem[Cherti et~al.(2023)Cherti, Beaumont, Wightman, Wortsman, Ilharco, Gordon, Schuhmann, Schmidt, and Jitsev]{cherti2023reproducible}
Mehdi Cherti, Romain Beaumont, Ross Wightman, Mitchell Wortsman, Gabriel Ilharco, Cade Gordon, Christoph Schuhmann, Ludwig Schmidt, and Jenia Jitsev.
\newblock Reproducible scaling laws for contrastive language-image learning.
\newblock In \emph{Proceedings of the IEEE/CVF conference on computer vision and pattern recognition}, pages 2818--2829, 2023.

\bibitem[Cho et~al.(2022)Cho, Yoon, Kale, Dernoncourt, Bui, and Bansal]{cho2022fine}
Jaemin Cho, Seunghyun Yoon, Ajinkya Kale, Franck Dernoncourt, Trung Bui, and Mohit Bansal.
\newblock Fine-grained image captioning with clip reward.
\newblock \emph{arXiv preprint arXiv:2205.13115}, 2022.

\bibitem[Cornia et~al.(2020)Cornia, Stefanini, Baraldi, and Cucchiara]{cornia2020meshed}
Marcella Cornia, Matteo Stefanini, Lorenzo Baraldi, and Rita Cucchiara.
\newblock Meshed-memory transformer for image captioning.
\newblock In \emph{Proceedings of the IEEE/CVF conference on computer vision and pattern recognition}, pages 10578--10587, 2020.

\bibitem[Costa et~al.(2007)Costa, Lorena, Carvalho, and Freitas]{costa2007review}
Eduardo Costa, Ana Lorena, ACPLF Carvalho, and Alex Freitas.
\newblock A review of performance evaluation measures for hierarchical classifiers.
\newblock In \emph{Evaluation methods for machine learning II: Papers from the AAAI-2007 workshop}, pages 1--6, 2007.

\bibitem[Dai et~al.(2023)Dai, Li, Li, Tiong, Zhao, Wang, Li, Fung, and Hoi]{instructblip}
Wenliang Dai, Junnan Li, Dongxu Li, Anthony Meng~Huat Tiong, Junqi Zhao, Weisheng Wang, Boyang Li, Pascale Fung, and Steven Hoi.
\newblock Instructblip: Towards general-purpose vision-language models with instruction tuning, 2023.

\bibitem[Deria et~al.(2025)Deria, Dukre, Tang, Atito, Roy, Awais, Khan, and Razzak]{deria2025dual}
Ankan Deria, Adinath~Madhavrao Dukre, Feilong Tang, Sara Atito, Sudipta Roy, Muhammad Awais, Muhammad~Haris Khan, and Imran Razzak.
\newblock Dual-stage value-guided inference with margin-based reward adjustment for fast and faithful vlm captioning.
\newblock \emph{arXiv preprint arXiv:2506.15649}, 2025.

\bibitem[Dess{\`\i} et~al.(2023)Dess{\`\i}, Bevilacqua, Gualdoni, Rakotonirina, Franzon, and Baroni]{dessi2023cross}
Roberto Dess{\`\i}, Michele Bevilacqua, Eleonora Gualdoni, Nathana{\"e}l~Carraz Rakotonirina, Francesca Franzon, and Marco Baroni.
\newblock Cross-domain image captioning with discriminative finetuning.
\newblock In \emph{Proceedings of the IEEE/CVF conference on computer vision and pattern recognition}, pages 6935--6944, 2023.

\bibitem[Dong et~al.(2024{\natexlab{a}})Dong, Li, Wu, Wang, Zhang, and Guo]{dong2024benchmarking}
Hongyuan Dong, Jiawen Li, Bohong Wu, Jiacong Wang, Yuan Zhang, and Haoyuan Guo.
\newblock Benchmarking and improving detail image caption.
\newblock \emph{arXiv preprint arXiv:2405.19092}, 2024{\natexlab{a}}.

\bibitem[Dong et~al.(2024{\natexlab{b}})Dong, Zhang, Zang, Cao, Wang, Ouyang, Wei, Zhang, Duan, Cao, et~al.]{dong2024internlm}
Xiaoyi Dong, Pan Zhang, Yuhang Zang, Yuhang Cao, Bin Wang, Linke Ouyang, Xilin Wei, Songyang Zhang, Haodong Duan, Maosong Cao, et~al.
\newblock Internlm-xcomposer2: Mastering free-form text-image composition and comprehension in vision-language large model.
\newblock \emph{arXiv preprint arXiv:2401.16420}, 2024{\natexlab{b}}.

\bibitem[Dong et~al.(2025)Dong, Dong, Wang, Huang, Zhou, Sun, Jing, Lan, Zhu, and Zheng]{dong2025inter}
Xin Dong, Shichao Dong, Jin Wang, Jing Huang, Li Zhou, Zenghui Sun, Lihua Jing, Jinsong Lan, Xiaoyong Zhu, and Bo Zheng.
\newblock Inter: Mitigating hallucination in large vision-language models by interaction guidance sampling.
\newblock In \emph{Proceedings of the IEEE/CVF International Conference on Computer Vision}, pages 2534--2544, 2025.

\bibitem[Duan et~al.(2025)Duan, Sun, Peng, Liu, Song, and Hu]{duan2025fuzzy}
Siyuan Duan, Yuan Sun, Dezhong Peng, Zheng Liu, Xiaomin Song, and Peng Hu.
\newblock Fuzzy multimodal learning for trusted cross-modal retrieval.
\newblock In \emph{Proceedings of the Computer Vision and Pattern Recognition Conference}, pages 20747--20756, 2025.

\bibitem[Dzabraev et~al.(2024)Dzabraev, Kunitsyn, and Ivaniuta]{dzabraev2024vlrm}
Maksim Dzabraev, Alexander Kunitsyn, and Andrei Ivaniuta.
\newblock Vlrm: Vision-language models act as reward models for image captioning.
\newblock \emph{arXiv preprint arXiv:2404.01911}, 2024.

\bibitem[et~al(2023)]{leng2023mitigating}
Leng et al.
\newblock Mitigating object hallucinations in large vision-language models through visual contrastive decoding.
\newblock \emph{arXiv preprint arXiv:2311.16922}, 2023.

\bibitem[Favero et~al.(2024)Favero, Zancato, Trager, Choudhary, Perera, Achille, Swaminathan, and Soatto]{favero2024multi}
Alessandro Favero, Luca Zancato, Matthew Trager, Siddharth Choudhary, Pramuditha Perera, Alessandro Achille, Ashwin Swaminathan, and Stefano Soatto.
\newblock Multi-modal hallucination control by visual information grounding.
\newblock In \emph{Proceedings of the IEEE/CVF Conference on Computer Vision and Pattern Recognition}, pages 14303--14312, 2024.

\bibitem[Fei et~al.(2023)Fei, Wang, Zhang, He, Wang, and Zheng]{fei2023transferable}
Junjie Fei, Teng Wang, Jinrui Zhang, Zhenyu He, Chengjie Wang, and Feng Zheng.
\newblock Transferable decoding with visual entities for zero-shot image captioning.
\newblock In \emph{Proceedings of the IEEE/CVF international conference on computer vision}, pages 3136--3146, 2023.

\bibitem[Feng et~al.(2019)Feng, Ma, Liu, and Luo]{feng2019unsupervised}
Yang Feng, Lin Ma, Wei Liu, and Jiebo Luo.
\newblock Unsupervised image captioning.
\newblock In \emph{Proceedings of the IEEE/CVF Conference on Computer Vision and Pattern Recognition}, pages 4125--4134, 2019.

\bibitem[Gaur et~al.(2024)Gaur, Singh, and Tapaswi]{gaur2024no}
Manu Gaur, Darshan Singh, and Makarand Tapaswi.
\newblock No detail left behind: Revisiting self-retrieval for fine-grained image captioning.
\newblock \emph{arXiv preprint arXiv:2409.03025}, 2024.

\bibitem[Hu et~al.(2024)Hu, Li, Lu, Shao, He, Qiao, and Luo]{hu2024omnimedvqa}
Yutao Hu, Tianbin Li, Quanfeng Lu, Wenqi Shao, Junjun He, Yu Qiao, and Ping Luo.
\newblock Omnimedvqa: A new large-scale comprehensive evaluation benchmark for medical lvlm.
\newblock In \emph{Proceedings of the IEEE/CVF Conference on Computer Vision and Pattern Recognition}, pages 22170--22183, 2024.

\bibitem[Huang et~al.(2019)Huang, Wang, Chen, and Wei]{huang2019attention}
Lun Huang, Wenmin Wang, Jie Chen, and Xiao-Yong Wei.
\newblock Attention on attention for image captioning.
\newblock In \emph{Proceedings of the IEEE/CVF international conference on computer vision}, pages 4634--4643, 2019.

\bibitem[Huang et~al.(2024)Huang, Dong, Zhang, Wang, He, Wang, Lin, Zhang, and Yu]{huang2024opera}
Qidong Huang, Xiaoyi Dong, Pan Zhang, Bin Wang, Conghui He, Jiaqi Wang, Dahua Lin, Weiming Zhang, and Nenghai Yu.
\newblock Opera: Alleviating hallucination in multi-modal large language models via over-trust penalty and retrospection-allocation.
\newblock In \emph{Proceedings of the IEEE/CVF Conference on Computer Vision and Pattern Recognition}, pages 13418--13427, 2024.

\bibitem[Jung et~al.(2025)Jung, Lee, Kim, and Yoon]{jung2025visual}
Mingi Jung, Saehyung Lee, Eunji Kim, and Sungroh Yoon.
\newblock Visual attention never fades: Selective progressive attention recalibration for detailed image captioning in multimodal large language models.
\newblock \emph{arXiv preprint arXiv:2502.01419}, 2025.

\bibitem[Kim et~al.(2025)Kim, Lee, Kim, and Kim]{kim2025vipcap}
Taewhan Kim, Soeun Lee, Si-Woo Kim, and Dong-Jin Kim.
\newblock Vipcap: Retrieval text-based visual prompts for lightweight image captioning.
\newblock In \emph{Proceedings of the AAAI Conference on Artificial Intelligence}, pages 4320--4328, 2025.

\bibitem[Kornblith et~al.(2023)Kornblith, Li, Wang, and Nguyen]{kornblith2023guiding}
Simon Kornblith, Lala Li, Zirui Wang, and Thao Nguyen.
\newblock Guiding image captioning models toward more specific captions.
\newblock In \emph{Proceedings of the IEEE/CVF International Conference on Computer Vision}, pages 15259--15269, 2023.

\bibitem[Lee et~al.(2025)Lee, Shin, Son, and Hwang]{lee2025diffusion}
Jeong~Ryong Lee, Yejee Shin, Geonhui Son, and Dosik Hwang.
\newblock Diffusion bridge: Leveraging diffusion model to reduce the modality gap between text and vision for zero-shot image captioning.
\newblock In \emph{Proceedings of the Computer Vision and Pattern Recognition Conference}, pages 4050--4059, 2025.

\bibitem[Lee et~al.(2024)Lee, Yoon, Bui, Shi, and Yoon]{lee2024toward}
Saehyung Lee, Seunghyun Yoon, Trung Bui, Jing Shi, and Sungroh Yoon.
\newblock Toward robust hyper-detailed image captioning: A multiagent approach and dual evaluation metrics for factuality and coverage.
\newblock \emph{arXiv preprint arXiv:2412.15484}, 2024.

\bibitem[Li et~al.(2024{\natexlab{a}})Li, Vo, Sugimoto, and Nakayama]{li2024evcap}
Jiaxuan Li, Duc~Minh Vo, Akihiro Sugimoto, and Hideki Nakayama.
\newblock Evcap: Retrieval-augmented image captioning with external visual-name memory for open-world comprehension.
\newblock In \emph{Proceedings of the IEEE/CVF conference on computer vision and pattern recognition}, pages 13733--13742, 2024{\natexlab{a}}.

\bibitem[Li et~al.(2024{\natexlab{b}})Li, Zhang, Diao, Wang, Wang, and Duan]{li2024densefusion}
Xiaotong Li, Fan Zhang, Haiwen Diao, Yueze Wang, Xinlong Wang, and Lingyu Duan.
\newblock Densefusion-1m: Merging vision experts for comprehensive multimodal perception.
\newblock \emph{Advances in Neural Information Processing Systems}, 37:\penalty0 18535--18556, 2024{\natexlab{b}}.

\bibitem[Lin et~al.(2025)Lin, Wang, Ren, and Han]{lin2025panoptic}
Kun-Yu Lin, Hongjun Wang, Weining Ren, and Kai Han.
\newblock Panoptic captioning: Seeking an equivalency bridge for image and text.
\newblock \emph{arXiv preprint arXiv:2505.16334}, 2025.

\bibitem[Lin et~al.(2014)Lin, Maire, Belongie, Hays, Perona, Ramanan, Doll{\'a}r, and Zitnick]{lin2014microsoft}
Tsung-Yi Lin, Michael Maire, Serge Belongie, James Hays, Pietro Perona, Deva Ramanan, Piotr Doll{\'a}r, and C~Lawrence Zitnick.
\newblock Microsoft coco: Common objects in context.
\newblock In \emph{European conference on computer vision}, pages 740--755. Springer, 2014.

\bibitem[Liu et~al.(2017{\natexlab{a}})Liu, Mao, Sha, and Yuille]{liu2017attention}
Chenxi Liu, Junhua Mao, Fei Sha, and Alan Yuille.
\newblock Attention correctness in neural image captioning.
\newblock In \emph{Proceedings of the AAAI conference on artificial intelligence}, 2017{\natexlab{a}}.

\bibitem[Liu et~al.(2023)Liu, Li, Li, and Lee]{liu2023improvedllava}
Haotian Liu, Chunyuan Li, Yuheng Li, and Yong~Jae Lee.
\newblock Improved baselines with visual instruction tuning, 2023.

\bibitem[Liu et~al.(2024)Liu, Li, Li, Li, Zhang, Shen, and Lee]{liu2024llavanext}
Haotian Liu, Chunyuan Li, Yuheng Li, Bo Li, Yuanhan Zhang, Sheng Shen, and Yong~Jae Lee.
\newblock Llava-next: Improved reasoning, ocr, and world knowledge, 2024.

\bibitem[Liu et~al.(2017{\natexlab{b}})Liu, Zhu, Ye, Guadarrama, and Murphy]{liu2017improved}
Siqi Liu, Zhenhai Zhu, Ning Ye, Sergio Guadarrama, and Kevin Murphy.
\newblock Improved image captioning via policy gradient optimization of spider.
\newblock In \emph{Proceedings of the IEEE international conference on computer vision}, pages 873--881, 2017{\natexlab{b}}.

\bibitem[Liu et~al.(2018)Liu, Li, Shao, Chen, and Wang]{liu2018show}
Xihui Liu, Hongsheng Li, Jing Shao, Dapeng Chen, and Xiaogang Wang.
\newblock Show, tell and discriminate: Image captioning by self-retrieval with partially labeled data.
\newblock In \emph{Proceedings of the European conference on computer vision (ECCV)}, pages 338--354, 2018.

\bibitem[Luo et~al.(2024)Luo, Chen, Li, Pan, Feng, Chao, and Yao]{luo2024unleashing}
Jianjie Luo, Jingwen Chen, Yehao Li, Yingwei Pan, Jianlin Feng, Hongyang Chao, and Ting Yao.
\newblock Unleashing text-to-image diffusion prior for zero-shot image captioning.
\newblock In \emph{European Conference on Computer Vision}, pages 237--254. Springer, 2024.

\bibitem[Luo et~al.(2023)Luo, Hu, Xi, Zhang, and Ma]{luo2023tuning}
Ziyang Luo, Zhipeng Hu, Yadong Xi, Rongsheng Zhang, and Jing Ma.
\newblock I-tuning: Tuning frozen language models with image for lightweight image captioning.
\newblock In \emph{ICASSP 2023-2023 IEEE International Conference on Acoustics, Speech and Signal Processing (ICASSP)}, pages 1--5. IEEE, 2023.

\bibitem[Mokady et~al.(2021)Mokady, Hertz, and Bermano]{mokady2021clipcap}
Ron Mokady, Amir Hertz, and Amit~H Bermano.
\newblock Clipcap: Clip prefix for image captioning.
\newblock \emph{arXiv preprint arXiv:2111.09734}, 2021.

\bibitem[Onoe et~al.(2024)Onoe, Rane, Berger, Bitton, Cho, Garg, Ku, Parekh, Pont-Tuset, Tanzer, et~al.]{onoe2024docci}
Yasumasa Onoe, Sunayana Rane, Zachary Berger, Yonatan Bitton, Jaemin Cho, Roopal Garg, Alexander Ku, Zarana Parekh, Jordi Pont-Tuset, Garrett Tanzer, et~al.
\newblock Docci: Descriptions of connected and contrasting images.
\newblock In \emph{European Conference on Computer Vision}, pages 291--309. Springer, 2024.

\bibitem[Park et~al.(2024)Park, Lee, Choe, and Chang]{park2024convis}
Yeji Park, Deokyeong Lee, Junsuk Choe, and Buru Chang.
\newblock Convis: Contrastive decoding with hallucination visualization for mitigating hallucinations in multimodal large language models.
\newblock \emph{arXiv preprint arXiv:2408.13906}, 2024.

\bibitem[Parkhi et~al.(2012)Parkhi, Vedaldi, Zisserman, and Jawahar]{parkhi2012cats}
Omkar~M Parkhi, Andrea Vedaldi, Andrew Zisserman, and CV Jawahar.
\newblock Cats and dogs.
\newblock In \emph{2012 IEEE conference on computer vision and pattern recognition}, pages 3498--3505. IEEE, 2012.

\bibitem[Peng et~al.(2025)Peng, He, Wei, Wen, and Hu]{peng2025patch}
Ruotian Peng, Haiying He, Yake Wei, Yandong Wen, and Di Hu.
\newblock Patch matters: Training-free fine-grained image caption enhancement via local perception.
\newblock In \emph{Proceedings of the Computer Vision and Pattern Recognition Conference}, pages 3963--3973, 2025.

\bibitem[Pi et~al.(2024)Pi, Zhang, Zhang, Pan, Chen, and Zhang]{pi2024image}
Renjie Pi, Jianshu Zhang, Jipeng Zhang, Rui Pan, Zhekai Chen, and Tong Zhang.
\newblock Image textualization: An automatic framework for creating accurate and detailed image descriptions.
\newblock \emph{arXiv preprint arXiv:2406.07502}, 2024.

\bibitem[Pont-Tuset et~al.(2020)Pont-Tuset, Uijlings, Changpinyo, Soricut, and Ferrari]{pont2020connecting}
Jordi Pont-Tuset, Jasper Uijlings, Soravit Changpinyo, Radu Soricut, and Vittorio Ferrari.
\newblock Connecting vision and language with localized narratives.
\newblock In \emph{European conference on computer vision}, pages 647--664. Springer, 2020.

\bibitem[Qiu et~al.(2025)Qiu, Gao, Qian, Pan, Yu, Li, Wang, Tang, Zhuang, and Chua]{qiu2025step}
Haiyi Qiu, Minghe Gao, Long Qian, Kaihang Pan, Qifan Yu, Juncheng Li, Wenjie Wang, Siliang Tang, Yueting Zhuang, and Tat-Seng Chua.
\newblock Step: Enhancing video-llms' compositional reasoning by spatio-temporal graph-guided self-training.
\newblock In \emph{Proceedings of the Computer Vision and Pattern Recognition Conference}, pages 3284--3294, 2025.

\bibitem[Qu et~al.(2024)Qu, Sun, Wei, and Cheng]{qu2024look}
Xiaoye Qu, Jiashuo Sun, Wei Wei, and Yu Cheng.
\newblock Look, compare, decide: Alleviating hallucination in large vision-language models via multi-view multi-path reasoning.
\newblock \emph{arXiv preprint arXiv:2408.17150}, 2024.

\bibitem[Radford et~al.(2021)Radford, Kim, Hallacy, Ramesh, Goh, Agarwal, Sastry, Askell, Mishkin, Clark, et~al.]{radford2021learning}
Alec Radford, Jong~Wook Kim, Chris Hallacy, Aditya Ramesh, Gabriel Goh, Sandhini Agarwal, Girish Sastry, Amanda Askell, Pamela Mishkin, Jack Clark, et~al.
\newblock Learning transferable visual models from natural language supervision.
\newblock In \emph{International conference on machine learning}, pages 8748--8763. PmLR, 2021.

\bibitem[Ramos et~al.(2023)Ramos, Martins, Elliott, and Kementchedjhieva]{ramos2023smallcap}
Rita Ramos, Bruno Martins, Desmond Elliott, and Yova Kementchedjhieva.
\newblock Smallcap: lightweight image captioning prompted with retrieval augmentation.
\newblock In \emph{Proceedings of the IEEE/CVF Conference on Computer Vision and Pattern Recognition}, pages 2840--2849, 2023.

\bibitem[Reimers and Gurevych(2019)]{reimers2019sentence}
Nils Reimers and Iryna Gurevych.
\newblock Sentence-bert: Sentence embeddings using siamese bert-networks.
\newblock \emph{arXiv preprint arXiv:1908.10084}, 2019.

\bibitem[Rotstein et~al.(2024)Rotstein, Bensaid, Brody, Ganz, and Kimmel]{rotstein2024fusecap}
Noam Rotstein, David Bensaid, Shaked Brody, Roy Ganz, and Ron Kimmel.
\newblock Fusecap: Leveraging large language models for enriched fused image captions.
\newblock In \emph{Proceedings of the IEEE/CVF winter conference on applications of computer vision}, pages 5689--5700, 2024.

\bibitem[Saito et~al.(2025)Saito, Kim, Park, Hashimoto, and Ushiku]{saito2025captionsmiths}
Kuniaki Saito, Donghyun Kim, Kwanyong Park, Atsushi Hashimoto, and Yoshitaka Ushiku.
\newblock Captionsmiths: Flexibly controlling language pattern in image captioning.
\newblock \emph{arXiv preprint arXiv:2507.01409}, 2025.

\bibitem[Saravanan et~al.(2025)Saravanan, Gupta, Singh, Khan, Gandhi, and Tapaswi]{saravanan2025velociti}
Darshana Saravanan, Varun Gupta, Darshan Singh, Zeeshan Khan, Vineet Gandhi, and Makarand Tapaswi.
\newblock Velociti: Benchmarking video-language compositional reasoning with strict entailment.
\newblock In \emph{Proceedings of the Computer Vision and Pattern Recognition Conference}, pages 18914--18924, 2025.

\bibitem[Shao et~al.(2024)Shao, Wang, Zhu, Xu, Song, Bi, Zhang, Zhang, Li, Wu, et~al.]{shao2024deepseekmath}
Zhihong Shao, Peiyi Wang, Qihao Zhu, Runxin Xu, Junxiao Song, Xiao Bi, Haowei Zhang, Mingchuan Zhang, YK Li, Yang Wu, et~al.
\newblock Deepseekmath: Pushing the limits of mathematical reasoning in open language models.
\newblock \emph{arXiv preprint arXiv:2402.03300}, 2024.

\bibitem[Sheng et~al.(2025)Sheng, Zhang, Ye, Wu, Zhang, Zhang, Peng, Lin, and Wu]{sheng2025hybridflow}
Guangming Sheng, Chi Zhang, Zilingfeng Ye, Xibin Wu, Wang Zhang, Ru Zhang, Yanghua Peng, Haibin Lin, and Chuan Wu.
\newblock Hybridflow: A flexible and efficient rlhf framework.
\newblock In \emph{Proceedings of the Twentieth European Conference on Computer Systems}, pages 1279--1297, 2025.

\bibitem[Sim{\'e}oni et~al.(2025)Sim{\'e}oni, Vo, Seitzer, Baldassarre, Oquab, Jose, Khalidov, Szafraniec, Yi, Ramamonjisoa, et~al.]{simeoni2025dinov3}
Oriane Sim{\'e}oni, Huy~V Vo, Maximilian Seitzer, Federico Baldassarre, Maxime Oquab, Cijo Jose, Vasil Khalidov, Marc Szafraniec, Seungeun Yi, Micha{\"e}l Ramamonjisoa, et~al.
\newblock Dinov3.
\newblock \emph{arXiv preprint arXiv:2508.10104}, 2025.

\bibitem[Song et~al.(2020)Song, Tan, Qin, Lu, and Liu]{song2020mpnet}
Kaitao Song, Xu Tan, Tao Qin, Jianfeng Lu, and Tie-Yan Liu.
\newblock Mpnet: Masked and permuted pre-training for language understanding.
\newblock \emph{Advances in neural information processing systems}, 33:\penalty0 16857--16867, 2020.

\bibitem[Tam et~al.(2023)Tam, Raffel, and Bansal]{tam2023simple}
Derek Tam, Colin Raffel, and Mohit Bansal.
\newblock Simple weakly-supervised image captioning via clip's multimodal embeddings.
\newblock In \emph{The AAAI-23 Workshop on Creative AI Across Modalities}, 2023.

\bibitem[Tewel et~al.(2022)Tewel, Shalev, Schwartz, and Wolf]{tewel2022zerocap}
Yoad Tewel, Yoav Shalev, Idan Schwartz, and Lior Wolf.
\newblock Zerocap: Zero-shot image-to-text generation for visual-semantic arithmetic.
\newblock In \emph{Proceedings of the IEEE/CVF conference on computer vision and pattern recognition}, pages 17918--17928, 2022.

\bibitem[Vinyals et~al.(2015)Vinyals, Toshev, Bengio, and Erhan]{vinyals2015show}
Oriol Vinyals, Alexander Toshev, Samy Bengio, and Dumitru Erhan.
\newblock Show and tell: A neural image caption generator.
\newblock In \emph{Proceedings of the IEEE conference on computer vision and pattern recognition}, pages 3156--3164, 2015.

\bibitem[Wang et~al.(2022)Wang, Yang, Hu, Li, Lin, Gan, Liu, Liu, and Wang]{wang2022git}
Jianfeng Wang, Zhengyuan Yang, Xiaowei Hu, Linjie Li, Kevin Lin, Zhe Gan, Zicheng Liu, Ce Liu, and Lijuan Wang.
\newblock Git: A generative image-to-text transformer for vision and language.
\newblock \emph{arXiv preprint arXiv:2205.14100}, 2022.

\bibitem[Wang et~al.(2024{\natexlab{a}})Wang, Dong, Zhu, Yao, Zhao, Li, and Luo]{wang2024diagnosing}
Jin Wang, Shichao Dong, Yapeng Zhu, Kelu Yao, Weidong Zhao, Chao Li, and Ping Luo.
\newblock Diagnosing the compositional knowledge of vision language models from a game-theoretic view.
\newblock \emph{arXiv preprint arXiv:2405.17201}, 2024{\natexlab{a}}.

\bibitem[Wang et~al.(2024{\natexlab{b}})Wang, Bai, Tan, Wang, Fan, Bai, Chen, Liu, Wang, Ge, et~al.]{wang2024qwen2}
Peng Wang, Shuai Bai, Sinan Tan, Shijie Wang, Zhihao Fan, Jinze Bai, Keqin Chen, Xuejing Liu, Jialin Wang, Wenbin Ge, et~al.
\newblock Qwen2-vl: Enhancing vision-language model's perception of the world at any resolution.
\newblock \emph{arXiv preprint arXiv:2409.12191}, 2024{\natexlab{b}}.

\bibitem[Wang et~al.(2020)Wang, Wei, Dong, Bao, Yang, and Zhou]{wang2020minilm}
Wenhui Wang, Furu Wei, Li Dong, Hangbo Bao, Nan Yang, and Ming Zhou.
\newblock Minilm: Deep self-attention distillation for task-agnostic compression of pre-trained transformers.
\newblock \emph{Advances in neural information processing systems}, 33:\penalty0 5776--5788, 2020.

\bibitem[Wang et~al.(2024{\natexlab{c}})Wang, Chen, Wang, Cao, Liu, Gao, Zhu, Zhu, Lu, Qiao, and Dai]{wang2024mpo}
Weiyun Wang, Zhe Chen, Wenhai Wang, Yue Cao, Yangzhou Liu, Zhangwei Gao, Jinguo Zhu, Xizhou Zhu, Lewei Lu, Yu Qiao, and Jifeng Dai.
\newblock Enhancing the reasoning ability of multimodal large language models via mixed preference optimization.
\newblock \emph{arXiv preprint arXiv:2411.10442}, 2024{\natexlab{c}}.

\bibitem[Xing et~al.(2025)Xing, Dong, Zang, Cao, Liang, Huang, Wang, Wu, and Lin]{xing2025caprl}
Long Xing, Xiaoyi Dong, Yuhang Zang, Yuhang Cao, Jianze Liang, Qidong Huang, Jiaqi Wang, Feng Wu, and Dahua Lin.
\newblock Caprl: Stimulating dense image caption capabilities via reinforcement learning.
\newblock \emph{arXiv preprint arXiv:2509.22647}, 2025.

\bibitem[Xu et~al.(2023)Xu, Zhao, Cai, and Huang]{xu2023zero}
Dongsheng Xu, Wenye Zhao, Yi Cai, and Qingbao Huang.
\newblock Zero-textcap: Zero-shot framework for text-based image captioning.
\newblock In \emph{Proceedings of the 31st ACM International Conference on Multimedia}, pages 4949--4957, 2023.

\bibitem[Yang et~al.(2025{\natexlab{a}})Yang, Li, Yang, Zhang, Hui, Zheng, Yu, Gao, Huang, Lv, et~al.]{yang2025qwen3}
An Yang, Anfeng Li, Baosong Yang, Beichen Zhang, Binyuan Hui, Bo Zheng, Bowen Yu, Chang Gao, Chengen Huang, Chenxu Lv, et~al.
\newblock Qwen3 technical report.
\newblock \emph{arXiv preprint arXiv:2505.09388}, 2025{\natexlab{a}}.

\bibitem[Yang et~al.(2025{\natexlab{b}})Yang, Feng, Yan, Wang, Wang, Zhu, Zhang, Xiao, Wu, Zhu, et~al.]{yang2025bacon}
Zhantao Yang, Ruili Feng, Keyu Yan, Huangji Wang, Zhicai Wang, Shangwen Zhu, Han Zhang, Jie Xiao, Pingyu Wu, Kai Zhu, et~al.
\newblock Bacon: Improving clarity of image captions via bag-of-concept graphs.
\newblock In \emph{Proceedings of the Computer Vision and Pattern Recognition Conference}, pages 14380--14389, 2025{\natexlab{b}}.

\bibitem[Ye et~al.(2023)Ye, Xu, Xu, Ye, Yan, Zhou, Wang, Hu, Shi, Shi, Li, Xu, Chen, Tian, Qi, Zhang, and Huang]{ye2023mplugowl}
Qinghao Ye, Haiyang Xu, Guohai Xu, Jiabo Ye, Ming Yan, Yiyang Zhou, Junyang Wang, Anwen Hu, Pengcheng Shi, Yaya Shi, Chenliang Li, Yuanhong Xu, Hehong Chen, Junfeng Tian, Qian Qi, Ji Zhang, and Fei Huang.
\newblock mplug-owl: Modularization empowers large language models with multimodality.
\newblock \emph{arXiv preprint arXiv:2304.14178}, 2023.

\bibitem[Yu et~al.(2023)Yu, Li, Hao, Zhu, Xu, and He]{yu2023cgt}
Jiarui Yu, Haoran Li, Yanbin Hao, Bin Zhu, Tong Xu, and Xiangnan He.
\newblock Cgt-gan: Clip-guided text gan for image captioning.
\newblock In \emph{Proceedings of the 31st ACM international conference on multimedia}, pages 2252--2263, 2023.

\bibitem[Yuksekgonul et~al.(2022)Yuksekgonul, Bianchi, Kalluri, Jurafsky, and Zou]{yuksekgonul2022and}
Mert Yuksekgonul, Federico Bianchi, Pratyusha Kalluri, Dan Jurafsky, and James Zou.
\newblock When and why vision-language models behave like bags-of-words, and what to do about it?
\newblock \emph{arXiv preprint arXiv:2210.01936}, 2022.

\bibitem[Zeng et~al.(2024)Zeng, Xie, Zhang, Chen, Chen, and Wang]{zeng2024meacap}
Zequn Zeng, Yan Xie, Hao Zhang, Chiyu Chen, Bo Chen, and Zhengjue Wang.
\newblock Meacap: Memory-augmented zero-shot image captioning.
\newblock In \emph{Proceedings of the IEEE/CVF conference on computer vision and pattern recognition}, pages 14100--14110, 2024.

\bibitem[Zhang et~al.(2025{\natexlab{a}})Zhang, Zeng, Li, Yu, and Chen]{zhang2025sc}
Lin Zhang, Xianfang Zeng, Kangcong Li, Gang Yu, and Tao Chen.
\newblock Sc-captioner: Improving image captioning with self-correction by reinforcement learning.
\newblock In \emph{Proceedings of the IEEE/CVF International Conference on Computer Vision}, pages 23145--23155, 2025{\natexlab{a}}.

\bibitem[Zhang et~al.(2025{\natexlab{b}})Zhang, Peng, Zhang, Guo, Wu, Huang, Chen, Ke, Meng, and Sun]{zhang2025will}
Xiaoying Zhang, Da Peng, Yipeng Zhang, Zonghao Guo, Chengyue Wu, Jen-Tse Huang, Chi Chen, Wei Ke, Helen Meng, and Maosong Sun.
\newblock Will pre-training ever end? a first step toward next-generation foundation mllms via self-improving systematic cognition.
\newblock \emph{arXiv preprint arXiv:2503.12303}, 2025{\natexlab{b}}.

\bibitem[Zhao et~al.(2025)Zhao, Zhang, Zhang, Wang, and Li]{zhao2025mitigating}
Fei Zhao, Chengcui Zhang, Runlin Zhang, Tianyang Wang, and Xi Li.
\newblock Mitigating image captioning hallucinations in vision-language models.
\newblock \emph{arXiv preprint arXiv:2505.03420}, 2025.

\bibitem[Zhu et~al.(2025)Zhu, Wang, Chen, Liu, Ye, Gu, Tian, Duan, Su, Shao, et~al.]{zhu2025internvl3}
Jinguo Zhu, Weiyun Wang, Zhe Chen, Zhaoyang Liu, Shenglong Ye, Lixin Gu, Hao Tian, Yuchen Duan, Weijie Su, Jie Shao, et~al.
\newblock Internvl3: Exploring advanced training and test-time recipes for open-source multimodal models.
\newblock \emph{arXiv preprint arXiv:2504.10479}, 2025.

\bibitem[Zou et~al.(2024)Zou, Wang, Yan, Huang, Zheng, Chen, Tang, and Hu]{zou2024look}
Xin Zou, Yizhou Wang, Yibo Yan, Sirui Huang, Kening Zheng, Junkai Chen, Chang Tang, and Xuming Hu.
\newblock Look twice before you answer: Memory-space visual retracing for hallucination mitigation in multimodal large language models.
\newblock \emph{arXiv preprint arXiv:2410.03577}, 2024.

\end{thebibliography}
}

\clearpage
\setcounter{page}{1}
\maketitlesupplementary

\section{Prompt Templates}
We adopt simple and concise prompts to perform image captioning.
For relation evaluation and hierarchical classification, we prompt Qwen3~\cite{yang2025qwen3} models to answer the given questions based on the candidate captions.
The prompts are detailed as follows:
\begin{figure}[h]
    \centering
    \includegraphics[width=\linewidth]{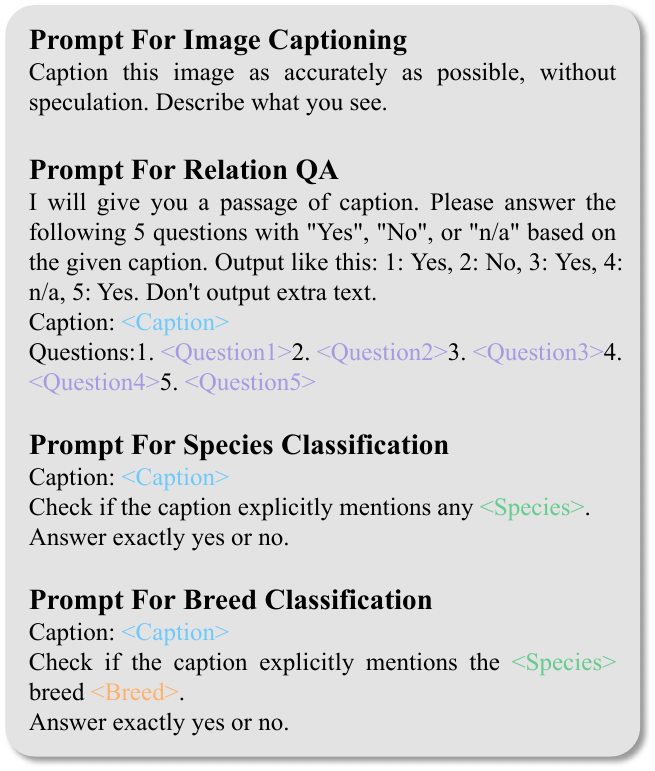}
    \caption{\textbf{Prompts for image captioning, relation evaluation, and hierarchical classification.}}
    \label{fig:prompt}
\end{figure}

\section{Additional Experiments}

\subsection{Comparisons with more Models}
We evaluate Qwen2-VL-7B~\cite{wang2024qwen2}, ShareCaptioner~\cite{zhang2025sc}, Gemini‑1.5, and Claude‑3.7 for image captioning and compare the results on the DOCCI500~\cite{onoe2024docci} benchmark, as shown in \cref{tab:more_models}.
We observe that closed‑source models (Gemini‑1.5 and Claude‑3.7) generally outperform the baseline Qwen2‑VL‑7B in CAPTURE, Object F1, Attribute F1, and Relation QA metrics.
However, after applying our proposed reinforcement learning method CIM to the Qwen2‑VL‑7B SFT model, the model (`SFT + Ours') achieves the best performance across all metrics, surpassing both open‑source and closed‑source models.
\begin{table}[h]
\resizebox{\columnwidth}{!}{%
\tablestyle{2pt}{1.5}
\centering
\begin{tabular}{lccccccc}
\hline
\multirow{2}{*}{Model} &
\multirow{2}{*}{BLEU-4} &
\multirow{2}{*}{METEOR} &
\multirow{2}{*}{CAPTURE} &
Objects & Attributes & Relations \\ \cline{5-7}
&&&& F1 & F1 & QA \\\hline
\vb Qwen2-VL-7B~\cite{wang2024qwen2} & 29.39 & 16.59 & 57.96 & 66.47 & 52.65 & 17.57 \\
\vb ShareCaptioner~\cite{zhang2025sc} & 39.09 & 23.05 & 57.90 & 66.05 & 52.27 & 19.47 \\
\vb Gemini-1.5 & 24.70 & 16.25 & 60.34 & 68.17 & 55.69 & 28.90 \\
\vb Claude-3.7 & 41.36 & 20.17 & 61.02 & 69.48 & 55.57 & 28.78 \\
\oursbg \cb SFT + Ours & 
\oursbg \textbf{45.16} &
\oursbg \textbf{24.88} &
\oursbg \textbf{64.31} &
\oursbg \textbf{73.87} &
\oursbg \textbf{58.68} &
\oursbg \textbf{36.32}
\\\hline
\end{tabular}
} 
\caption{\textbf{Performance comparison of open-source and closed-source models on the DOCCI500~\cite{onoe2024docci} benchmark.} Best results are highlighted in bold. Our model achieves the best performance across all metrics, surpassing both open‑source and closed‑source models.}
\label{tab:more_models}
\end{table}

\subsection{Comparisons with Larger Models}
As shown in \cref{tab:larger_models}, reinforcement learning with CIM consistently boosts the performance of 7B/8B models, enabling them to surpass much larger models (72B/76B/78B) on both COCO-LN500~\cite{pont2020connecting} and DOCCI500~\cite{onoe2024docci}. 
Averaged over all backbones and benchmarks, our models achieve improvements of 2.3\%, 3.6\%, 2.0\%, 4.3\%, 5.0\%, and 5.1\% on BLEU-4, METEOR, CAPTURE, Objects F1, Attributes F1, and Relation QA compared to their larger counterparts.
These results demonstrate that our CIM-based reinforcement learning framework has substantially unlocked the potential of 7B/8B models, enabling them to generate more detailed and precise captions than their much larger counterparts.

\begin{table*}[!ht]
\tablestyle{2.0pt}{1.72}
    \centering
    \begin{tabular}{lllcccccccccc}
    \hline
       \multirow{2}{*}{\makecell{Base Model}} &  
       \multirow{2}{*}{Benchmark} &  
       \multirow{2}{*}{Method} & 
       \multirow{2}{*}{BLEU-4} &
       \multirow{2}{*}{METEOR} &
       \multirow{2}{*}{CAPTURE} & 
       \multicolumn{3}{c}{Objects} & 
       \multicolumn{3}{c}{Attributes} & 
       Relations \\
        \cline{7-13}
        &&&&&& Precision & Recall & F1 & Precision & Recall & F1 & QA \\ \hline
       \multirow{6}{*}{\makecell{Qwen2-VL~\cite{wang2024qwen2}}} & 
       \multirow{3}{*}{\makecell{COCO-LN500}} & 
       \vb 7B-Base & 39.57 & 20.42 & 46.52 & 81.12 & 61.82 & 69.47 & 66.48 & 42.86 & 48.68 & 20.47 \\
       && \vb 72B-Base & 39.37 & 19.77 & 47.20 & \textbf{82.52} & 62.89 & 70.72 & 69.36 & 42.78 & 49.38 & 27.46 \\
       && \oursbg \cb 7B-Ours & \oursbg \textbf{39.63} & \oursbg \textbf{27.06} & \oursbg \textbf{48.64} & \oursbg 80.90 & \oursbg \textbf{72.23} & \oursbg \textbf{75.80} & \oursbg \textbf{72.45} & \oursbg \textbf{54.08} & \oursbg \textbf{58.22} & \oursbg \textbf{38.71} \\ 
       \cline{2-13}
       & 
       \multirow{3}{*}{\makecell{DOCCI500}} & 
       \vb 7B-Base & 29.39 & 16.59 & 57.96 & 83.69 & 56.79 & 66.47 & 69.96 & 43.27 & 52.65 & 17.57 \\
       && \vb 72B-Base & 30.79 & 17.57 & 59.56 & \textbf{84.02} & 60.21 & 69.14 & 71.63 & 44.31 & 53.92 & 27.21 \\
       && \oursbg \cb 7B-Ours & \oursbg \textbf{37.21} & \oursbg \textbf{21.53} & \oursbg \textbf{63.12} & \oursbg 81.99 & \oursbg \textbf{64.61} & \oursbg \textbf{71.43} & \oursbg \textbf{74.49} & \oursbg \textbf{50.19} & \oursbg \textbf{59.18} & \oursbg \textbf{32.12} \\ \hline

       \multirow{6}{*}{\makecell{Qwen2.5-VL~\cite{bai2025qwen2}}} & 
       \multirow{3}{*}{\makecell{COCO-LN500}} & 
       \vb 7B-Base & 29.11 & 14.58 & 44.12 & \textbf{82.35} & 55.72 & 65.37 & 66.30 & 39.90 & 46.25 & 23.76 \\ 
       && \vb 72B-Base & \textbf{40.21} & 22.57 & 48.06 & 81.21 & 66.37 & 72.42 & 68.46 & 48.21 & 52.97 & 32.13 \\ 
       && \oursbg \cb 7B-Ours & \oursbg 37.15 & \oursbg \textbf{26.81} & \oursbg \textbf{48.93} & \oursbg 79.28 & \oursbg \textbf{75.91} & \oursbg \textbf{77.59} & \oursbg \textbf{72.49} & \oursbg \textbf{54.46} & \oursbg \textbf{58.51} & \oursbg \textbf{44.15} \\ 
       \cline{2-13}
       & 
       \multirow{3}{*}{\makecell{DOCCI500}} & 
       \vb 7B-Base & 22.68 & 14.67 & 55.89 & \textbf{84.64} & 54.96 & 65.06 & 72.15 & 42.13 & 52.27 & 24.35 \\ 
       && \vb 72B-Base & 33.72 & 19.01 & 61.09 & 82.17 & 62.92 & 70.35 & 72.37 & 46.18 & 55.46 & 30.71 \\ 
       && \oursbg \cb 7B-Ours & \oursbg \textbf{39.80} & \oursbg \textbf{22.72} & \oursbg \textbf{63.46} & \oursbg 79.28 & \oursbg \textbf{66.97} & \oursbg \textbf{71.88} & \oursbg \textbf{73.52} & \oursbg \textbf{50.45} & \oursbg \textbf{59.08} & \oursbg \textbf{34.70} \\ \hline

       \multirow{6}{*}{\makecell{InternVL2~\cite{chen2024internvl}}} & 
       \multirow{3}{*}{\makecell{COCO-LN500}} & 
       \vb 8B-Base & 31.44 & 21.28 & 45.86 & 80.46 & 65.20 & 71.05 & 70.63 & 47.40 & 52.73 & 26.65 \\ 
       && \vb 76B-Base & 37.31 & 22.29 & 47.64 & 80.57 & 67.25 & 72.66 & 70.69 & 46.90 & 54.50 & 31.60 \\ 
       && \oursbg \cb 8B-Ours & \oursbg \textbf{41.89} & \oursbg \textbf{25.85} & \oursbg \textbf{49.06} & \oursbg \textbf{82.05} & \oursbg \textbf{69.40} & \oursbg \textbf{74.57} & \oursbg \textbf{73.50} & \oursbg \textbf{52.59} & \oursbg \textbf{57.49} & \oursbg \textbf{35.54} \\ 
       \cline{2-13}
       & 
       \multirow{3}{*}{\makecell{DOCCI500}} & 
       \vb 8B-Base & 31.49 & 17.97 & 58.83 & 81.49 & 59.54 & 67.72 & 70.84 & 44.24 & 53.51 & 22.65 \\ 
       && \vb 76B-Base & \textbf{38.69} & \textbf{21.21} & \textbf{61.09} & 81.42 & \textbf{64.77} & \textbf{71.19} & 70.90 & 46.80 & 55.57 & \textbf{29.67} \\ 
       && \oursbg \cb 8B-Ours & \oursbg 32.01 & \oursbg 19.49 & \oursbg 60.82 & \oursbg \textbf{82.16} & \oursbg 60.35 & \oursbg 68.66 & \oursbg \textbf{74.16} & \oursbg \textbf{47.51} & \oursbg \textbf{56.98} & \oursbg 26.36 \\ \hline

       \multirow{6}{*}{\makecell{InternVL2.5~\cite{chen2024expanding,wang2024mpo}}} & 
       \multirow{3}{*}{\makecell{COCO-LN500}} & 
       \vb 8B-Base & 38.04 & 18.54 & 47.09 & \textbf{82.42} & 64.02 & 71.42 & \textbf{72.00} & 46.03 & 52.50 & 27.29 \\ 
       && \vb 78B-Base & \textbf{41.22} & 20.79 & 48.25 & 81.38 & 64.59 & 71.42 & 71.25 & 46.84 & 52.85 & 28.92 \\ 
       && \oursbg \cb 8B-Ours & \oursbg 39.27 & \oursbg \textbf{22.85} & \oursbg \textbf{48.28} & \oursbg 80.31 & \oursbg \textbf{70.42} & \oursbg \textbf{74.53} & \oursbg 71.85 & \oursbg \textbf{51.71} & \oursbg \textbf{56.07} & \oursbg \textbf{36.96} \\ 
       \cline{2-13}
       & \multirow{3}{*}{\makecell{DOCCI500}} & 
       \vb 8B-Base & 24.23 & 15.29 & 58.64 & 84.11 & 58.07 & 67.62 & 73.48 & 44.79 & 54.69 & 24.63 \\ 
       && \vb 78B-Base & 23.60 & 15.79 & 59.37 & \textbf{85.78} & 59.12 & 68.80 & \textbf{75.12} & 45.50 & 55.74 & 26.92 \\ 
       && \oursbg \cb 8B-Ours & \oursbg \textbf{30.89} & \oursbg \textbf{17.84} & \oursbg \textbf{60.75} & \oursbg 82.17 & \oursbg \textbf{62.28} & \oursbg \textbf{69.99} & \oursbg 72.96 & \oursbg \textbf{46.64} & \oursbg \textbf{56.07} & \oursbg \textbf{28.90} \\ \hline

       \multirow{6}{*}{\makecell{InternVL3~\cite{zhu2025internvl3}}} & 
       \multirow{3}{*}{\makecell{COCO-LN500}} & 
       \vb 8B-Base & 33.21 & 16.16 & 47.88 & \textbf{82.51} & 63.31 & 71.00 & 71.14 & 43.97 & 50.66 & 26.44 \\ 
       && \vb 78B-Base & 40.60 & 19.50 & 48.48 & 82.46 & 65.02 & 72.07 & 71.76 & 46.22 & 52.50 & 29.61 \\ 
       && \oursbg \cb 8B-Ours & \oursbg \textbf{40.64} & \oursbg \textbf{25.64} & \oursbg \textbf{48.90} & \oursbg 80.33 & \oursbg \textbf{73.33} & \oursbg \textbf{76.14} & \oursbg \textbf{73.33} & \oursbg \textbf{54.54} & \oursbg \textbf{58.70} & \oursbg \textbf{38.67} \\ 
       \cline{2-13}
       & 
       \multirow{3}{*}{\makecell{DOCCI500}} & 
       \vb 8B-Base & 12.57 & 11.66 & 56.95 & 85.98 & 55.39 & 66.08 & 74.75 & 43.19 & 53.72 & 25.11 \\ 
       && \vb 78B-Base & 18.32 & 13.89 & 58.52 & \textbf{86.18} & 58.23 & 68.30 & 75.69 & 44.33 & 55.01 & 27.85 \\ 
       && \oursbg \cb 8B-Ours & \oursbg \textbf{28.25} & \oursbg \textbf{18.54} & \oursbg \textbf{62.18} & \oursbg 82.73 & \oursbg \textbf{62.86} & \oursbg \textbf{70.47} & \oursbg \textbf{76.22} & \oursbg \textbf{49.58} & \oursbg \textbf{59.26} & \oursbg \textbf{30.39} \\ \hline
    \end{tabular}

    \caption{
    \textbf{Performance of Reinforcement Learning with CIM on Base Models 
    over COCO-LN500~\cite{pont2020connecting} and DOCCI500~\cite{onoe2024docci}.} 
    Best results are highlighted in bold.
    The results indicate that post-training the base model with our CIM enhances its ability to generate high-quality captions, and even 7B/8B models can outperform much larger baselines.
  }
        \label{tab:larger_models}
\end{table*}

\subsection{Training on the DOCCI Training Set}
We also use the training set of DOCCI~\cite{onoe2024docci} which consists of 9.7K image-caption pairs as the training set for supervised fine-tuning and reinforcement learning of Qwen2-VL-7B~\cite{wang2024qwen2}. 
Evaluation results on the COCO-LN500~\cite{pont2020connecting} and DOCCI500~\cite{onoe2024docci} benchmarks are reported in \cref{tab:docci}.
Our proposed method consistently outperforms both the SFT model and SC-Captioner~\cite{zhang2025sc} by a substantial margin.
For instance, compared with SC-Captioner, our models achieve improvements of 1.2\% in Attributes F1 on COCO-LN500 and 1.3\% on DOCCI500.
Similarly, it achieves improvements of 4.8\% in Relations QA on COCO-LN500 and 7.3\% on DOCCI500.
These consistent gains across both datasets highlight the robustness and generalizability of our reinforcement learning approach.

In the COCO-LN500~\cite{pont2020connecting} setting, the ground-truth captions are generally shorter, which leads to relatively low precision scores. This occurs because shorter captions often omit certain objects and details present in the images, thereby penalizing the precision metric even when these elements are correctly identified by the model. Similarly, the highest BLEU-4 score observed for the base model can be attributed to this characteristic of the dataset.

\begin{table*}[!htbp]
\tablestyle{2.0pt}{1.5}
\centering
\begin{tabular}{lllccccccccccc}
\hline
\multirow{2}{*}{\makecell{Base Model}} &
\multirow{2}{*}{Benchmark} &
\multirow{2}{*}{Method} &
\multirow{2}{*}{BLEU-4} &
\multirow{2}{*}{METEOR} &
\multirow{2}{*}{CAPTURE} &
\multicolumn{3}{c}{Objects} &
\multicolumn{3}{c}{Attributes} &
Relations \\
\cline{7-13}
&&&&&& Precision & Recall & F1 & Precision & Recall & F1 & QA \\ \hline

\multirow{8}{*}{\makecell{Qwen2-VL-7B~\cite{wang2024qwen2}}} &
\multirow{4}{*}{\makecell{COCO-LN500}} &
\vb Base & \textbf{39.57} & 20.42 & 46.52 & \textbf{81.12} & 61.82 & 69.47 & 66.48 & 42.86 & 48.68 & 20.47 \\
&& \vb SFT & 34.93 & 25.97 & 47.63 & 78.03 & 70.42 & 73.43 & 67.91 & 53.51 & 56.31 & 30.06 \\
&& \vb SC-Captioner~\cite{zhang2025sc}   & 32.75 & 26.17 & 47.92 & 78.58 & 72.09 & 74.60 & 69.01 & 54.03 & 56.93 & 32.01 \\
&& \oursbg \cb SFT + Ours & \oursbg 32.34 & \oursbg \textbf{26.50} & \oursbg \textbf{48.36} & \oursbg 78.76 & \oursbg \textbf{72.93} & \oursbg \textbf{75.13} & \oursbg \textbf{69.72} & \oursbg \textbf{55.23} & \oursbg \textbf{58.10} & \oursbg \textbf{36.76} \\
\cline{2-13}
& \multirow{4}{*}{\makecell{DOCCI500}} &
\vb Base & 29.39 & 16.59 & 57.96 & \textbf{83.69} & 56.79 & 66.47 & 69.96 & 43.27 & 52.65 & 17.57 \\
&& \vb SFT & 40.29 & 25.44 & 62.53 & 78.01 & 65.30 & 70.31 & 67.13 & 49.33 & 56.08 & 25.43 \\
&& \vb SC-Captioner~\cite{zhang2025sc}   & 41.95 & 26.38 & 63.83 & 78.85 & 67.59 & 72.05 & 69.77 & 50.64 & 58.00 & 28.58 \\
&& \oursbg \cb SFT + Ours & \oursbg \textbf{42.62} & \oursbg \textbf{27.13} & \oursbg \textbf{64.48} & \oursbg 80.03 & \oursbg \textbf{68.59} & \oursbg \textbf{73.10} & \oursbg \textbf{71.83} & \oursbg \textbf{51.58} & \oursbg \textbf{59.30} & \oursbg \textbf{35.91} \\\hline
\end{tabular}
\caption{\textbf{Performance of Reinforcement Learning with CIM on the SFT Model, trained on the DOCCI~\cite{onoe2024docci} training set, over COCO-LN500~\cite{pont2020connecting} and DOCCI500~\cite{onoe2024docci}.} Best results are highlighted in bold. Our model outperforms both the SFT model and SC-Captioner~\cite{zhang2025sc}.}
\label{tab:docci}
\end{table*}

\section{Visualization Examples}
\subsection{Comparisons with the Base Model}
\begin{figure*}[h]
    \centering
    \includegraphics[width=\linewidth]{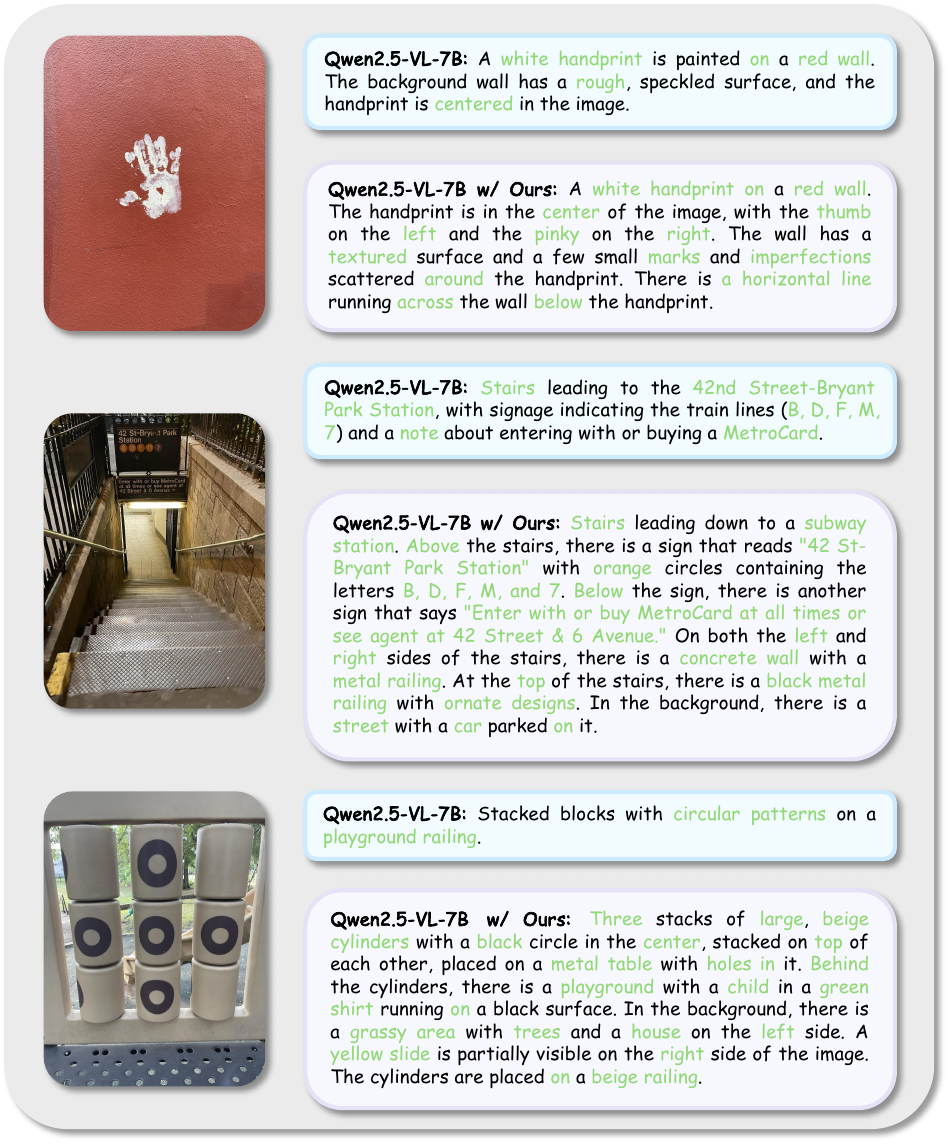}
    \caption{\textbf{Qualitative comparison between the base model and the base model trained with reinforcement learning using CIM.} 
    Green annotations indicate correct descriptions.
    Our method enhances the base model’s ability to produce more detailed captions.}
    \label{fig:case_ours}
\end{figure*}

To directly illustrate the effect of reinforcement learning with CIM on the base model, we present qualitative examples comparing the original base model and the base model trained with CIM, as shown in \cref{fig:case_ours}.
We mark correct descriptions in green for clarity. 
As can be observed, captions generated by the base model often overlook fine-grained details, whereas our method is able to generate more comprehensive descriptions, demonstrating improved descriptive capability.
\subsection{Comparisons with the SFT Model}
\begin{figure*}[h]
    \centering
    \includegraphics[width=\linewidth]{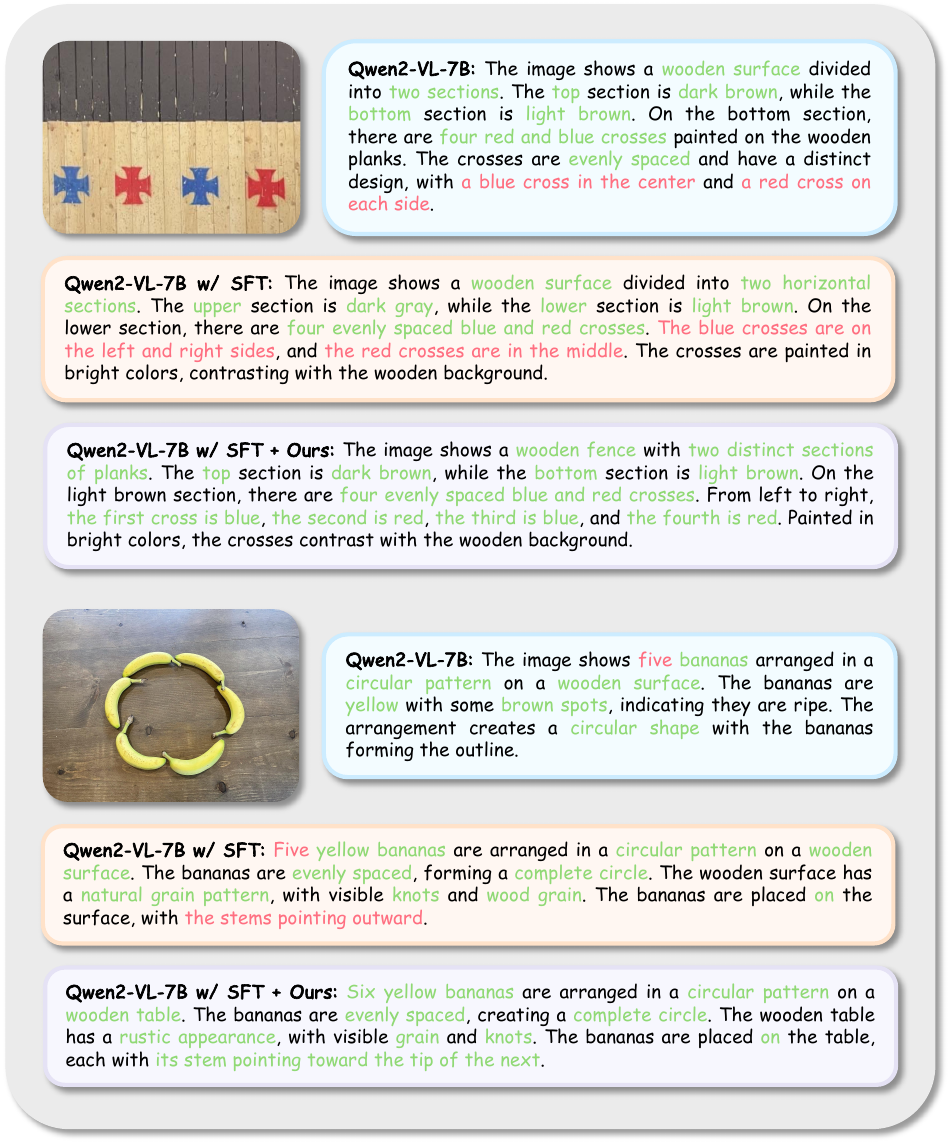}
    \caption{\textbf{Qualitative comparison among the base model, the SFT model, and the SFT model trained with reinforcement learning using CIM.} 
    Green annotations indicate correct descriptions, while red annotations indicate incorrect descriptions. 
    Compared with the base model and the SFT model, our method produces more detailed captions while maintaining the precision of caption.}
    \label{fig:case_sft_ours}
\end{figure*}

To further validate the effectiveness of our approach, we provide a qualitative comparison among the base model, the SFT model, and the SFT model trained with reinforcement learning using CIM, as shown in Fig.~\ref{fig:case_sft_ours}. 
Correct descriptions are marked in green and incorrect descriptions in red. 
It can be observed that captions generated by the base model often miss fine-grained details. 
In contrast, our proposed method can generate more detailed descriptions while maintaining caption precision compared with SFT model.

\end{document}